\documentclass[10pt]{article} %
\usepackage[accepted]{tmlr}

\usepackage{amsmath,amsfonts,bm}

\def\eqref#1{(\ref{#1})}

\def\1{\bm{1}}

\DeclareMathAlphabet{\mathsfit}{\encodingdefault}{\sfdefault}{m}{sl}
\SetMathAlphabet{\mathsfit}{bold}{\encodingdefault}{\sfdefault}{bx}{n}

\DeclareMathOperator*{\argmax}{arg\,max}

\usepackage{hyperref}
\usepackage{url}

\usepackage{algorithm}

\let\classAND\AND
\let\AND\relax
\usepackage{algorithmic}

\let\AND\classAND
\AtBeginEnvironment{algorithmic}{\let\AND\algoAND}

\usepackage{graphicx}
\usepackage{subcaption}
\usepackage{booktabs} %
\usepackage{wrapfig}
\usepackage{multirow}
\usepackage{tikz}

\usepackage{amsmath}
\usepackage{amssymb}
\usepackage{mathtools}
\usepackage{amsthm}

\theoremstyle{plain}
\newtheorem{theorem}{Theorem}[section]

\newtheorem{corollary}[theorem]{Corollary}
\theoremstyle{definition}
\newtheorem{definition}[theorem]{Definition}

\newtheorem{remark}[theorem]{Remark}

\title{Leveraging Task Structures for Improved Identifiability in Neural Network Representations}

\author{\name Wenlin Chen\thanks{Equal contribution.} \email wc337@cam.ac.uk \\
      \addr University of Cambridge, Cambridge, United Kingdom \\
      Max Planck Institute for Intelligent Systems, Tübingen, Germany
      \AND
      \name Julien Horwood$^*$ \email jdh95@cam.ac.uk \\
      \addr University of Cambridge, Cambridge, United Kingdom
      \AND
      \name Juyeon Heo \email jh2324@cam.ac.uk\\
      \addr  University of Cambridge, Cambridge, United Kingdom
      \AND
      \name José Miguel Hernández-Lobato \email jmh233@cam.ac.uk\\
      \addr  University of Cambridge, Cambridge, United Kingdom}

\def\month{08}  %
\def\year{2024} %
\def\openreview{\url{https://openreview.net/forum?id=WLcPrq6pu0}} %

\begin{document}

\maketitle

\begin{abstract}
This work extends the theory of identifiability in supervised learning by considering the consequences of having access to a distribution of tasks.
In such cases, we show that linear identifiability is achievable in the general multi-task regression setting.
Furthermore, we show that the existence of a task distribution which defines a conditional prior over latent factors reduces the equivalence class for identifiability to permutations and scaling of the true latent factors, a stronger and more useful result than linear identifiability. Crucially, when we further assume a causal structure over these tasks, our approach enables simple maximum marginal likelihood optimization, and suggests potential downstream applications to causal representation learning. Empirically, we find that this straightforward optimization procedure enables our model to outperform more general unsupervised models in recovering canonical representations for both synthetic data and real-world molecular data.
\end{abstract}

\section{Introduction}

Multi-task regression is a common problem in machine learning, which naturally arises in many scientific applications such as molecular property prediction \citep{stanley2021fsmol,chen2023metalearning} and machine learning force fields \citep{jacobson2023leveraging}. Despite this, most deep learning approaches to
this problem attempt to model the relationships between tasks through heuristic approaches, such as fitting a shared neural network in an attempt to capture the joint structures between tasks.
Beyond lacking a principled approach to modeling task relationships, these approaches fail to account for how we may expect the latent factors for related tasks to change. In this work, we show that by leveraging assumptions about the relationships between the latent factors of the data \textit{across} tasks, in particular that they vary in their causal and spurious relationships with the target variables, we can achieve identifiability of the latent factors up to permutations and scaling. 

A common assumption in the causal representation learning literature, known as the sparse mechanism shift hypothesis \citep{scholkopf2019causality,scholkopf2021toward,perry2022causal}, states that changes across tasks arise from sparse changes in the underlying causal mechanisms. While we do not operate directly on structural causal models, our result arises by similarly considering the implications of sparse changes in the causal graph defining a multi-task learning setting. We accomplish this by first
extending the theory of identifiability in supervised learning to the multi-task regression setting for identifiability up to linear transformations (i.e., weak identifiability). We then propose a new approach to identifying neural network representations up to permutations and scaling (i.e., strictly strong identifiability), by leveraging the causal structures 
of the underlying latent factors for each task. 
We empirically validate our model's ability to recover the ground-truth latent structure of the data both in simulated settings where data is generated from our model and for real-world molecular data. This contrasts with current state-of-the-art approaches \textcolor{black}{such as \citep{Khemakhem_Kingma_Monti_Hyvarinen_2020,Lu_Wu_Hernandez-Lobato_Scholkopf_2022}, whose assumptions also fit our assumed data generating process but which are difficult to train effectively and only identifiable up to block permutations and scaling of the sufficient statistics of their exponential family priors.} We summarize our contributions in the following section.

\subsection{Contributions}
Our work extends the current identifiability literature in the following key aspects.
\begin{enumerate}
    \item In contrast to prior work \citep{Lachapelle_Deleu_Mahajan_Mitliagkas_Bengio_Lacoste-Julien_Bertrand_2023,Fumero_Wenzel_Zancato_Achille_Rodolà_Soatto_Schölkopf_Locatello_2023} which relates meta/multi-task learning to identifiability via explicit sparsity constraints, this work expands these conceptual connections \emph{beyond sparsity constraints} by considering the shared causal structure between tasks. This \emph{significantly} reduces the number of tasks needed to recover the true representations.
    \item Our method extends previous identifiability results by resolving the \emph{point-wise indeterminacies} of prior work \citep{Khemakhem_Kingma_Monti_Hyvarinen_2020,Lu_Wu_Hernandez-Lobato_Scholkopf_2022}.
    \item Our model extends the applicability of conditional prior models to discriminative settings at test time as our identifiability result does \emph{not} require conditioning on the target values during inference.
    \item To our knowledge, our approach is the first to propose a conditionally \emph{factorized} prior model which can achieve identifiability via optimizing the \emph{exact} marginal likelihood. This leads to improved empirical results in our experiments despite the probabilistic assumptions of our model.
    \item While many works have shown that spurious correlations are a failure case of deep learning and focus on eliminating them \citep{rojas2018invariant,arjovsky2019invariant,krueger2021out,eastwood2022probable,Lu_Wu_Hernandez-Lobato_Scholkopf_2022,kirichenko2023last}, we leverage spurious features to \emph{improve} the robustness of learned representations in the multi-task setting through our identifiability results.
\end{enumerate}

\subsection{Organization of the Paper}
Section \ref{sec:related-work} reviews prior works that are closely or broadly related to identifiable and disentangled representation learning.
Section \ref{sec:proposed_method} describes the proposed identifiable multi-task representation learning method and discusses model assumptions, theoretical identifiability guarantees, and potential limitations. 
Section \ref{sec:experiments} empirically evaluates the proposed method on both synthetic datasets and real-world molecular datasets. Section \ref{sec:conclusion} summarizes the paper and its potential impact.

\section{Related Work}\label{sec:related-work}
This section discusses prior work in the identifiable and disentangled representation learning literature.
\subsection{Disentanglement and ICA}
The notion of optimizing for disentangled representations gained traction in the recent unsupervised deep learning literature when it was proposed that this objective may be sufficient to improve desirable attributes such as interpretability, robustness, and generalization \citep{bengio2013representation,Higgins_Matthey_Pal_Burgess_Glorot_Botvinick_Mohamed_Lerchner_2017,chen2016infogan}. However, 
the notion of disentanglement alone is not intrinsically well-defined, as there may be many disentangled representations of the data which are seemingly equally valid. Thus it is not clear a priori that this criterion is sufficient to achieve the above desiderata \citep{Locatello_Bauer_Lucic_Raetsch_Gelly_Scholkopf_Bachem_2019}. In the identifiable representation learning literature, the \textit{correct} disentangled representation is assumed to be the one 
corresponding to the ground-truth data generating process. Thus, what is required is an \textit{identifiable} representation, which must be equivalent to the causal one for sufficiently expressive model classes. In the linear case, identifiability results exist 
in the classical literature for ICA, which requires non-Gaussianity assumptions on the sources for the data \citep{Herault_Jutten_1986,Comon_1994}. 

\subsection{Conditional Prior Models for Non-Linear ICA}
Many extensions of ICA to the non-linear case have been proposed, together with significant theoretical advances. In particular, \citet{hyvarinen2019nonlinear} extend this by assuming a conditionally factorized prior over the latent factors given some observed auxiliary variables, and propose a contrastive learning objective for recovering the inverse of the function which generated the observations.
iVAE \citep{Khemakhem_Kingma_Monti_Hyvarinen_2020} further extends this to the setting of noisy observations, drawing connections with variational autoencoders \citep{kingma2013auto} and enabling direct optimization via a variational objective. \citet{Lachapelle_López_Sharma_Everett_Priol_Lacoste_Lacoste-Julien_2022} demonstrate that strong identifiability results remain achievable under weaker conditions on the sufficient statistics of the prior \textit{if} the data generating process implies that the latent factors are governed by sparse mechanism shifts.
iCaRL \citep{Lu_Wu_Hernandez-Lobato_Scholkopf_2022} derives analogous results for the case where the prior over the latent factors is a more general non-factorized exponential family distribution. However, the complex nature of the non-factorized prior in iCaRL requires score matching, which is difficult to optimize in practice. \citet{khemakhem2020ice} explore general conditions for identifiability in energy-based models, and introduce the notion of linear identifiability. This is expanded upon in the context of classification models in \citet{Roeder_Metz_Kingma_2021}, showing that the representations obtained via the final hidden layer of a neural network may be identifiable up to \textit{linear} transformations when conditioning on the label set.
The works of \citet{halva2021disentangling, morioka2021independent} both obtain strong identifiability results by exploiting specific temporal or spatial structure in the encoded latents and modelling the joint distributions as dynamical systems, however their models do not translate well to the static setting, and their identifiability results remain restricted to non-linear coordinate-wise transformations of the latent variables. 
While \citet{Hyvärinen_Pajunen_1999, Khemakhem_Kingma_Monti_Hyvarinen_2020} show that identifiability is not achievable without any form of conditioning in the prior, \citet{Willetts_Paige_2022,Kivva_Rajendran_Ravikumar_Aragam_2022} recently extended the results in unsupervised generative models to the case of models with mixture model priors. This can be seen as providing analogous identifiability results to prior methods using conditionally factorized priors, without assumptions on the observability or the dimensionality of the conditioning variable. Nonetheless, these results do not apply to the exact likelihood, and it remains unclear to what extent the practical consistency and identifiability is achievable when optimizing a variational surrogate objective. 

\subsection{Structural Approaches to Identifiability}
In contrast, \citet{Brady_Zimmermann_Sharma_Schölkopf_von_Kugelgen_Brendel_2023} discuss identifiability results which arise from assumptions on the structure of the mixing function, specifically targeting dual objectives of compositionality with respect to partitions of the latent factors and invertibility of the mixing function. Thus, no distributional assumptions are made on the prior. While this approach has similarities with our proposal by introducing assumptions on how partitions of the latent space evolve with respect to well-defined objects, we propose a general setting which is not restricted to representation learning in visual scenes. Furthermore, by formalizing these assumptions within our probabilistic model, we eliminate the need for auxiliary terms in our optimization objective. 

Recent work \citep{Lachapelle_Deleu_Mahajan_Mitliagkas_Bengio_Lacoste-Julien_Bertrand_2023, Fumero_Wenzel_Zancato_Achille_Rodolà_Soatto_Schölkopf_Locatello_2023} has expanded this area of research to consider the multi-task and meta-learning settings, and thus investigate the connections between identifiability and the structure of the learning problem itself. However, their approach to achieving permutation-identifiable representations relies on introducing heuristic sparsity constraints, such as entropy and $L_2$-norm regularizers, within a bi-level optimization objective, which turns out to be difficult to solve both in theory and in practice \citep{sinha2017review}. In addition, their approaches are less applicable in practice since a huge number of tasks are required (more than $10^5$ in their experiments). This contrasts with the straightforward, principled and task-efficient optimization objective arising from our probabilistic model.

\section{Identifiable Multi-task Representation Learning}\label{sec:proposed_method}

We propose a novel method that leverages task structures in the multi-task regression setting to identify the ground-truth data representations up to permutations and scaling.

\subsection{Problem Formulation}
\begin{wrapfigure}{r}{0.25\columnwidth}
    \centering
    \scalebox{1.0}{
    \begin{tikzpicture}[>=latex]
        \node[draw, circle, fill=gray!30] (T) at (2.5,-0.5) {$t$};
        \node[draw, circle, fill=gray!30] (X) at (0.5,-0.5) {$\mathbf{x}$};
        \node[draw, circle] (Z_c) at (1.5,0.5) {$\mathbf{z}_c$};
        \node[draw, circle] (Z_s) at (1.5,-1.5) {$\mathbf{z}_s$};
        \node[draw, circle, fill=gray!30] (Y) at (1.5,-0.5) {$y$};
    
        \draw[->] (T) -- (Z_s);
        \draw[->] (T) -- (Z_c);
        \draw[->] (T) -- (Y);
        \draw[->] (Z_s) -- (X);
        \draw[->] (Z_c) -- (X);
        \draw[->] (Z_c) -- (Y);
        \draw[->] (Y) -- (Z_s);
    \end{tikzpicture}
    }
    \caption{Assumed data generating process.}
    \label{fig:causal-graph}
\end{wrapfigure}
The assumptions of the ground-truth data generating process considered in this paper are encapsulated in the causal graph shown in Figure \ref{fig:causal-graph}, where the input variable $\mathbf{x}\in\mathcal{X}\subseteq\mathbb{R}^n$, the target variable\footnote{Without loss of generality, we assume that $\mathbb{E}(y)=0$. This can be achieved by standardizing $y$ in practice.} $y\in\mathbb{R}$ and the task index variable $t\in\{1,\cdots,N_t\}$ are observed variables, and the latent factors $\mathbf{z}\in\mathbb{R}^d~(d\leq n)$ are unobserved variables. We assume that $\mathbf{x}$ is generated by transforming some (unobserved) ground-truth latent factors $\mathbf{z}^*$ with some unknown injective mixing function $\mathbf{f}_*:\mathbb{R}^d\to\mathcal{X}$, i.e., $\mathbf{x}=\mathbf{f}_*(\mathbf{z^*})$. 
To incorporate the sparse mechanism shift hypothesis across tasks, we further assume that each task $t$ has its own partition of the ground-truth latent factors $\mathbf{z}^*=\mathbf{z}_c^*\cup\mathbf{z}_s^*$ into a set of causal latent factors $\mathbf{z}_c^*$ and a set of spurious latent factors $\mathbf{z}_s^*$, and such partitions potentially vary across tasks. The target variable is assumed to be a weighted sum of the causal latent factors, i.e., $y=(\mathbf{w}_t^*)^{\text{T}}\mathbf{z}^*$, where $\mathbf{w}_t^*\in\mathbb{R}^d$ are the ground-truth regression weights for task $t$ which assign zero weights for the spurious latent factors. Note that there may be latent factors that are uncorrelated with $y$ in some
tasks, which can be included within $\mathbf{z}_c^*$ but with zero regression weights. The spurious latent factors are assumed to be generated from the target variable with a different linear correlation function in each task $t$. Our goal is to recover the unobserved ground truth latent factors $\mathbf{z}^*$ given an empirical task distribution $p(t)$ over $N_t$ training tasks and an empirical data distribution $p(\mathbf{x},y|t)$ for each task $t\in\{1,\cdots,N_t\}$. 

\begin{figure}[t]
    \centering
    \includegraphics[width=\linewidth]{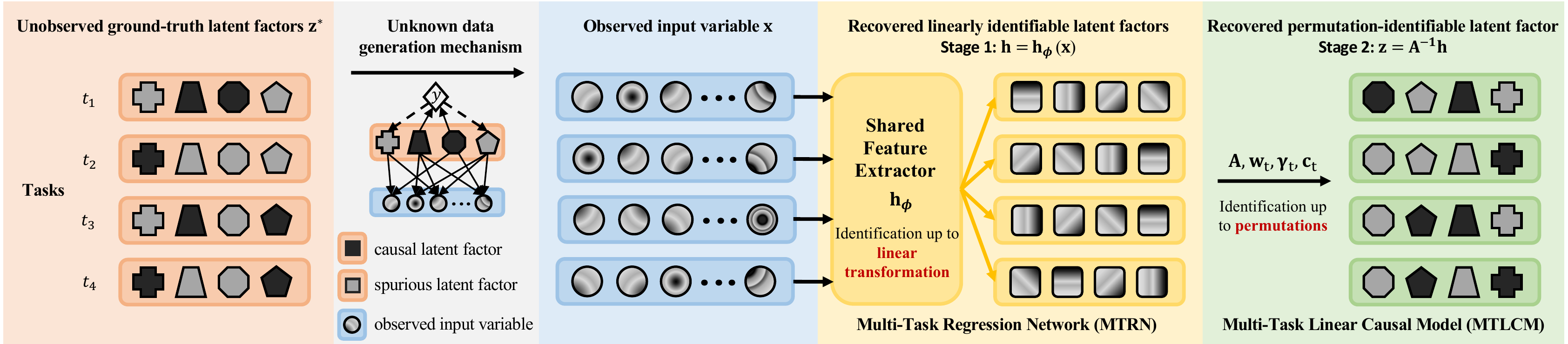}
    \caption{\textcolor{black}{The workflow of our proposed method. Shapes are used to track the positions of the ground-truth and recovered latent factors. Colors are used to differentiate between causal and spurious latent factors. We assume that the observed variable is obtained by transforming the ground-truth latent factors with some mixing function. We show that a multi-task regression network (MTRN) can recover the ground-truth latent factors (i.e., data representations) up to linear transformation and further propose a multi-task linear causal model (MTLCM) to reduce the equivalence class for identifiability to permutations and scaling.}}
    \label{fig:illustrative_figure}
\end{figure}

Overall, our proposed method consists of two stages as illustrated in Figure \ref{fig:illustrative_figure}. In the first stage (yellow), we train a multi-task regression network (MTRN) with a feature extractor shared across tasks and $N_t$ task-specific linear heads using maximum likelihood estimation (MLE). We show that upon convergence, the representations learned by the feature extractor are identifiable up to some invertible linear transformation (Corollary \ref{cor:lin-id-mtrn}). In the second stage (green), we use the assumed causal structure across tasks to define a conditional prior over the underlying independent latent factors. We show that this multi-task linear causal model (MTLCM) enables simple maximum marginal likelihood learning for recovering the linear transformation in the representations obtained in the first stage, which reduces the identifiability class to permutations and scaling (Theorem \ref{thm:strong-id}), and automatically disentangles and identifies the causes and effects of the target variable from the learned representations. 

\subsection{Stage 1: Multi-Task Regression Network}\label{sec:stage-1}
In the first stage, we train a multi-task regression network (MTRN) to recover the ground-truth latent factors up to some invertible linear transformation. 

Let $f_{\boldsymbol{\phi},\mathbf{w}_t}(\mathbf{x})=\mathbf{w}_t^{\text{T}}\mathbf{h}_{\boldsymbol{\phi}}(\mathbf{x})$ be the output of an MTRN for task $t$, 
where $\mathbf{w}_t\in\mathbb{R}^{d}$ are the regression weights in the linear head for task $t$, and $\mathbf{h}_{\boldsymbol{\phi}}(\mathbf{x})\in\mathbb{R}^{d}$ is the data representation produced by the feature extractor $\mathbf{h}_{\boldsymbol{\phi}}$ shared across all tasks with learnable parameters $\boldsymbol{\phi}$.
As in typical non-linear regression settings, the likelihood is assumed to be Gaussian $p_{\boldsymbol{\theta}}(y|\mathbf{x},t)=\mathcal{N}(y|f_{\boldsymbol{\phi},\mathbf{w}_t}(\mathbf{x}),\sigma_{r,t}^2)$ with mean modeled by an MTRN and variance fixed to some constant $\sigma_{r,t}^2$, where $\boldsymbol{\theta}\coloneqq(\boldsymbol{\phi},\mathbf{w}_1,\cdots,\mathbf{w}_{N_t})$ denotes all parameters in the MTRN.

We first define linearly identifiable (or weakly identifiable) representations in the multi-task setting.
\begin{definition}[Multi-task weak identifiability]\label{def:weak-id}
    Let $\boldsymbol{\theta}$ and $\boldsymbol{\theta}'$ be any two sets of parameters. Then, the data representations are \textit{linearly identifiable} if there exists an invertible matrix $\mathbf{A}\in\mathbb{R}^{d\times d}$ such that
    \begin{equation}
        p_{\boldsymbol{\theta}}(y|\mathbf{x},t)=p_{\boldsymbol{\theta}'}(y|\mathbf{x},t),~\forall t,\mathbf{x},y 
        \quad\implies\quad
        \mathbf{h}_{\boldsymbol{\phi}}(\mathbf{x})=\mathbf{A}\mathbf{h}_{\boldsymbol{\phi}'}(\mathbf{x}).
    \end{equation}
\end{definition}
We show that data representations of MTRN are linearly identifiable if we have access to a set of sufficiently diverse tasks measured by the linear dependencies among their regression weights. 
\begin{theorem}\label{thm:lin-id}
    Let $\boldsymbol{\theta}\coloneqq(\boldsymbol{\phi},\{\mathbf{w}_{t_i}\}_{i=1}^{N_t})$ and $\boldsymbol{\theta}'\coloneqq(\boldsymbol{\phi}',\{\mathbf{w}_{t_i}'\}_{i=1}^{N_t})$ be any two sets of parameters such that
    \begin{align}
        p_{\boldsymbol{\theta}}(y|\mathbf{x},t)=p_{\boldsymbol{\theta}'}(y|\mathbf{x},t),\quad\forall t,\mathbf{x},y.
    \end{align}
    Assume that $\text{Span}(\text{Im}(\mathbf{h}_{\boldsymbol{\phi}}))=\mathbb{R}^d$, i.e., the vectors in the image of the feature extractor $\mathbf{h}_{\boldsymbol{\phi}}$ span the whole $\mathbb{R}^d$. Suppose that there exist $d$ tasks $\{t_i\}_{i=1}^d\subseteq\{1,\cdots,N_t\}$ such that at least one set of regression weights (i.e., either $\{\mathbf{w}_{t_i}\}_{i=1}^d$ or $\{\mathbf{w}_{t_i}'\}_{i=1}^d$) are linearly independent.
    Then, the data representations of the MTRN are linearly identifiable.
\end{theorem}
The proof of Theorem \ref{thm:lin-id} can be found in Appendix \ref{appendix:proof-lin-id}.
Following standard practice, we train the MTRN via maximum likelihood estimation (MLE):
\begin{align}\label{eq:mtrn-loss} 
\boldsymbol{\theta}'=\argmax_{\boldsymbol{\theta}}~\mathbb{E}_{p(t)p(\mathbf{x},y|t)}[\log p_{\boldsymbol{\theta}}(y|\mathbf{x},t)].
\end{align}
Using Theorem \ref{thm:lin-id}, it is straightforward to show that MTRN trained with maximum likelihood estimation (MLE) can recover the ground-truth data representations up to some invertible linear transformation.

\begin{corollary}\label{cor:lin-id-mtrn}
    Let $\mathbf{h}_{*}:\mathcal{X}\to\mathbb{R}^d$ be the ground-truth mapping from input variables to the ground-truth latent factors, i.e., $\mathbf{z}^*=\mathbf{h}_{*}(\mathbf{x})$. Assume that $\text{Span}(\text{Im}(\mathbf{h}_{*}))=\mathbb{R}^d$. Suppose that there exist $d$ tasks $\{t_i\}_{i=1}^d\subseteq\{1,\cdots,N_t\}$ such that the set of ground-truth regression weights $\{\mathbf{w}_{t_i}^*\}_{i=1}^d$ are linearly independent. Assume that \eqref{eq:mtrn-loss} has a unique solution. Suppose that the optimization procedure for \eqref{eq:mtrn-loss} converges to the optimal predictive likelihood under standard regularity conditions for MLE estimators \textcolor{black}{\citep{gurland1954regularity}}, i.e.,
\begin{align}
    p_{\boldsymbol{\theta}'}(y|\mathbf{x},t)=p_{*}(y|\mathbf{x},t)\coloneqq\mathcal{N}(y|(\mathbf{w}_t^*)^{\text{T}}\mathbf{h}_{*}(\mathbf{x}),\sigma_{r,t}^2),\quad\forall t,\mathbf{x},y.
\end{align}
Then, the feature extractor $\mathbf{h}_{\boldsymbol{\phi}'}$ is guaranteed to recover the ground-truth latent factors up to some invertible linear transformation $\mathbf{A}_*$, i.e., $\mathbf{h}_{\boldsymbol{\phi}'}(\mathbf{x})=\mathbf{A}_*\mathbf{h}_{*}(\mathbf{x})$.
\end{corollary}
Corollary \ref{cor:lin-id-mtrn} essentially states that the effective number of tasks defined by the number of independent ground-truth linear heads at least needs to be the same as the number of latent dimensions to guarantee multi-task linear identifiability.
\begin{remark}
    While \citet{Lachapelle_Deleu_Mahajan_Mitliagkas_Bengio_Lacoste-Julien_Bertrand_2023}[Proposition 2.2] prove a similar proposition on MLE invariance to linear feature transformations, their proposition is built upon their Assumption 2.1 that the learned feature extractor $\mathbf{h}_{\boldsymbol{\phi}'}$ is linearly equivalent to the ground truth feature extractor $\mathbf{h}_{*}$. However, they do not specify under what conditions this assumption will hold for the MLE objective; they only specify conditions for the \textcolor{black}{bi-level} objective with a sparsity regularizer in their Section 3. In contrast, our Corollary \ref{cor:lin-id-mtrn} explicitly reveals such conditions for MLE, i.e., $\text{Span}(\text{Im}(\mathbf{h}_{*}))=\mathbb{R}^d$ and \textcolor{black}{the existence of $d$ linearly independent ground-truth task-specific regression weight vectors $\{\mathbf{w}_{t_i}^*\}_{i=1}^d$}.
\end{remark}

\subsection{Stage 2: Multi-Task Linear Causal Model} 
In the second stage, we freeze the feature extractor $\mathbf{h}_{\boldsymbol{\phi}'}$ learned in the first stage and denote its representations by $\mathbf{h}\coloneqq\mathbf{h}_{\boldsymbol{\phi}'}(\mathbf{x})$. Corollary \ref{cor:lin-id-mtrn} suggests that $\mathbf{h}=\mathbf{A}_{*}\mathbf{z}^*$ for some invertible matrix $\mathbf{A}_{*}$. 
We propose a \textit{multi-task linear causal model} (MTLCM) to recover the ground-truth latent factors up permutations and scaling from $\mathbf{h}$ based on our assumed causal graph in Figure \ref{fig:causal-graph}. The core idea of the MTLCM is to model the change in the causal and spurious latent factors across tasks with learnable task-specific parameters.

Let $\mathcal{T}(t)=\{\mathbf{c}_t,\boldsymbol{\gamma}_t\}$ be a collection of task-specific variables associated with task $t$, which are free parameters to be learned from data, where $\mathbf{c}_t\in\{0,1\}^d$ are the causal indicator variables which determine the partition of $\mathbf{z}=\mathbf{z}_c\cup\mathbf{z}_s$ for the given task $t$ (i.e., $c_{t,i}=1$ indicates that $z_i$ is a causal latent factor in task $t$ and $c_{t,i}=0$ indicates that $z_i$ is a spurious latent factor in task $t$), and $\boldsymbol{\gamma}_t$ are the coefficients used to generate the spurious latents from $y$ for task $t$. 

\subsubsection{Conditionally Factorized Prior Given Task and Target Variables}
Following the standard setting of generative models, the prior distribution over causal latent factors $\mathbf{z}_c$ are assumed to be a standard Gaussian distribution:
\begin{align}
    p_{\mathcal{T}}(\mathbf{z}_c|t)=\mathcal{N}(\mathbf{z}_c|\mathbf{0},\mathbf{I}),\label{eq:causal_prior}
\end{align}
which depends on the task variable $t$ since the causal indicator variable $\mathbf{c}_t$ that determines which subset of latent factors are causal varies across tasks.

According to the assumed data generating process, the target $y$ is a linear function of the latent data representations $\mathbf{z}$. Following the common setting of the last layer in a non-linear regression neural network, we assume that $y$ is generated from $\mathbf{z}_c$ via a linear Gaussian model with the regression weights $\mathbf{w}_t$ masked by the causal indicators $\mathbf{c}_t$:
\begin{align}\label{eq: regression-model}
    p_{\mathcal{T}}(y|\mathbf{z}_c,t) & =\mathcal{N}(y|(\mathbf{w}_t\circ\mathbf{c}_t)^{\text{T}}\mathbf{z},\sigma_{p}^{2}),
\end{align}
and that the spurious latent factors $\mathbf{z}_s$ are generated from $y$ via another linear Gaussian model:
\begin{align}
    p_{\mathcal{T}}(\mathbf{z}_s|y,t)=\mathcal{N}(\mathbf{z}_s|y\boldsymbol{\gamma}_t,\sigma_s^2\mathbf{I}).\label{eq:spurious_prior}
\end{align}
The structured conditional prior over all latent factors given $t$ and $y$ can then be obtained by Bayes' Rule:
\begin{equation}
    p_{\mathcal{T}}(\mathbf{z}|y,t)=\frac{p_{\mathcal{T}}(\mathbf{z}_c|t)p_{\mathcal{T}}(y|\mathbf{z}_c,t)p_{\mathcal{T}}(\mathbf{z}_s|y,t)}{\int p_{\mathcal{T}}(\mathbf{z}_c|t)p_{\mathcal{T}}(y|\mathbf{z}_c,t)p_{\mathcal{T}}(\mathbf{z}_s|y,t)d\mathbf{z}_s d\mathbf{z}_c}.
\end{equation}
Since no prior knowledge of regression weights $\mathbf{w}_t$ is assumed, we marginalize out $\mathbf{w}_t$ from $p_{\mathcal{T}}(y|\mathbf{z}_c,t)$ under an uninformative prior (i.e., an infinite-variance Gaussian prior). This makes the structured conditional prior factorize over all latent factors (see Appendix \ref{appendix:uninformative prior} for a derivation):
\begin{align}\label{eq:lin-causal-prior}
        p_{\mathcal{T}}(\mathbf{z}|y,t)
        =p_{\mathcal{T}}(\mathbf{z}_c|t)p_{\mathcal{T}}(\mathbf{z}_s|y,t)
        =\mathcal{N}(\mathbf{z}|\mathbf{a}_t,\boldsymbol{\Lambda}_t),
\end{align}
where the mean $\mathbf{a}_t$ and covariance $\boldsymbol{\Lambda}_t$ can be compactly expressed as:
\begin{equation}
    \mathbf{a}_t\coloneqq y\boldsymbol{\gamma}_t\circ(1-\mathbf{c}_t),\quad
    \boldsymbol{\Lambda}_t\coloneqq \text{diag}(\sigma_{s}^{2}(1-\mathbf{c}_t)+\mathbf{c}_t).\label{eq:con_prior_mean_var}
\end{equation}

\subsubsection{Linear Gaussian Likelihood}
Since the data representation $\mathbf{h}$ learned in the first stage is equivalent to $\mathbf{z}^*$ up to some linear transformation, we assume a linear Gaussian likelihood with invertible linear transformation $\mathbf{A}$, similar to the likelihood function in a probabilistic PCA model \citep{tipping1999probabilistic}:
\begin{align}\label{eq:s2-lkl}
    p_{\mathbf{A}}(\mathbf{h}|\mathbf{z}) & =\mathcal{N}(\mathbf{h}|\mathbf{A}\mathbf{z},\sigma_{o}^{2}\mathbf{I}),
\end{align}
where $\mathbf{A}$ is to be learned from data, which aims to recover the ground-truth linear transformation $\mathbf{A}_*$ for the linearly identifiable representation $\mathbf{h}$.

\subsubsection{Exact Maximum Marginal Likelihood Learning}
Let $\boldsymbol{\psi}=(\mathbf{A},\mathcal{T})$ denote all parameters in an MTLCM, including the linear transformation $\mathbf{A}$ and the task-specific parameters $\mathcal{T}(t)=\{\mathbf{c}_t,\boldsymbol{\gamma}_t\}$ for all tasks $t$. \textcolor{black}{The marginal likelihood for MTLCM is given by}
\begin{equation}\label{eq:s2-mlkl}
    p_{\boldsymbol{\psi}}(\mathbf{h}|y,t)=\int p_{\mathbf{A}}(\mathbf{h}|\mathbf{z})p_{\mathcal{T}}(\mathbf{z}|y,t)d\mathbf{z}
    =\mathcal{N}(\mathbf{h}|\boldsymbol{\mu}_t,\boldsymbol{\Sigma}_t),
\end{equation}
where the mean $\boldsymbol{\mu}_t$ and covariance $\boldsymbol{\Sigma}_t$ have closed-form expressions \textcolor{black}{(see Appendix \ref{appendix:derivation-mlkl} for a derivation)}:
\begin{equation}\label{eq:mtlcm-mlkl-mean-var}
\begin{split}
    \boldsymbol{\mu}_t = y\mathbf{A}(\boldsymbol{\gamma}_t\circ(1-\mathbf{c}_t)),\quad
    \boldsymbol{\Sigma}_t = \mathbf{A}\text{diag}(\sigma_{s}^{2}(1-\mathbf{c}_t)+\mathbf{c}_t)\mathbf{A}^{\text{T}}+\sigma_o^2\mathbf{I}.
\end{split}
\end{equation}
\begin{remark}
    The conditional prior $p(\mathbf{z}|y)$ over the latent factors $\mathbf{z}$ is typically non-factorized according to the data generating process described in Section \ref{sec:proposed_method}, since the causal latent factors $\mathbf{z}_c$ are parents of the target variable $y$, which become correlated when conditioning on $y$. 
    In order to guarantee strong identifiability, iCaRL \citep{Lu_Wu_Hernandez-Lobato_Scholkopf_2022} parameterizes such non-factorized conditional priors using ReLU activated energy-based models optimized by variational inference and score matching, which turns out to be difficult to train in practice due to variational overpruning \citep{trippe2018overpruning} and high computational complexity \citep{hyvarinen2005estimation}. In contrast, our proposed structured conditional prior \eqref{eq:lin-causal-prior} factorize over all latent factors, which, together with the linear Gaussian likelihood \eqref{eq:s2-lkl}, allows us to use exact maximum marginal likelihood learning for \eqref{eq:s2-mlkl} to recover the ground-truth latent factors $\mathbf{z}^*$ up to permutations and scaling from the linearly identifiable data representations $\mathbf{h}=\mathbf{h}_{\boldsymbol{\phi}}(\mathbf{x})$ learned in the first stage:
\begin{align}\label{eq:mtlcm-loss}
    \boldsymbol{\psi}'=\argmax_{\boldsymbol{\psi}}~\mathbb{E}_{p(t)p(\mathbf{x},y|t)}[\log p_{\boldsymbol{\psi}}(\mathbf{h}_{\boldsymbol{\phi}}(\mathbf{x})|y,t)].
\end{align}
\end{remark}
\begin{remark}
    It is worth noting that our method has greater applicability \textcolor{black}{for supervised learning} than the methods that rely on a learned probabilistic inverse $q_{\boldsymbol{\psi}}(\mathbf{z}|\mathbf{u})$ to extract identifiable latent factors from data such as iVAE \citep{Khemakhem_Kingma_Monti_Hyvarinen_2020} and iCaRL \citep{Lu_Wu_Hernandez-Lobato_Scholkopf_2022}. \textcolor{black}{While these approaches theoretically could apply to learned representations in discriminative settings by letting $\mathbf{u} = (\mathbf{x}, y)$, they are impractical in such contexts} since $q_{\boldsymbol{\psi}}(\mathbf{z}|\mathbf{x},y)$ depends on the target variable $y$ which is generally unknown at test time. In contrast, our method does not depend on $y$ at inference time, since the identifiable latent factors can be obtained by applying the inverse linear transformation learned by the MTLCM to the linearly identifiable data representations produced by the MTRN, i.e., $\mathbf{z}=\mathbf{A}^{-1}\mathbf{h}_{\boldsymbol{\phi}}(\mathbf{x})$. This enables our model to be applicable to discriminative settings at test time.
\end{remark}

\subsubsection{Identifiability Theory}
We first define the concept of strictly strong identifiability in the multi-task setting.
\begin{definition}[Strictly strong identifiability]\label{def:strong-id}
    Let $\boldsymbol{\psi}$ and $\boldsymbol{\psi}'$ be any two sets of parameters. The latent factors are identifiable up to permutations and scaling if there exists a permutation and scaling matrix $\mathbf{P}\in\mathbb{R}^{d\times d}$ 
    such that 
    \begin{equation}\label{eq:strong-id}
            p_{\boldsymbol{\psi}'}(\mathbf{h}|y,t)=p_{\boldsymbol{\psi}}(\mathbf{h}|y,t),~\forall \mathbf{h},t,y
            \quad\implies\quad
        (\mathbf{A')}^{-1}\mathbf{h}=\mathbf{P}\mathbf{A}^{-1}\mathbf{h}.
    \end{equation}
\end{definition}

We show that the latent factors of MTLCM are  strictly strongly identifiable if there are sufficient variations of causal/spurious latent factors across tasks measured by the linear dependencies among the natural parameters of their conditional priors.

\begin{theorem}\label{thm:strong-id}
    Let $\mathbf{u}\coloneqq[y,t]$ denote the conditioning variable and $k\coloneqq2d$. Assume that the learned and ground-truth linear transformations $\mathbf{A}$ and $\mathbf{A}_{*}$ are invertible. Suppose that there exist $k+1$ points $\mathbf{u}_0, \mathbf{u}_1,\cdots,\mathbf{u}_{k}$ such that
    \begin{align}
        \mathbf{L}\coloneqq[\boldsymbol{\eta}(\mathbf{u}_1)-\boldsymbol{\eta}(\mathbf{u}_0),\cdots,\boldsymbol{\eta}(\mathbf{u}_{k})-\boldsymbol{\eta}(\mathbf{u}_0)]
    \end{align}
    is invertible, where $\boldsymbol{\eta}(\mathbf{u})\coloneqq\left[\begin{smallmatrix}\boldsymbol{\Lambda}_t^{-1}\mathbf{a}_t \\ -\frac{1}{2}\text{diag}(\boldsymbol{\Lambda}_t)\end{smallmatrix}\right]\in\mathbb{R}^{k}$ are the natural parameters of $p_{\mathcal{T}}(\mathbf{z}|\mathbf{u})$. Assume that \eqref{eq:mtlcm-loss} has a unique solution. Suppose that the optimization procedure for \eqref{eq:mtlcm-loss} converges to the optimal marginal likelihood under standard regularity conditions for maximum marginal likelihood estimators \textcolor{black}{\citep{gurland1954regularity}}, i.e., for all $\mathbf{h},y,t$,
    \begin{equation}
        p_{\boldsymbol{\psi}'}(\mathbf{h}|y,t)=p_{*}(\mathbf{h}|y,t)\coloneqq\mathcal{N}(\mathbf{h}|\boldsymbol{\mu}_t^*,\boldsymbol{\Sigma}_t^*),
    \end{equation}
    where $\boldsymbol{\mu}_t^*$ and $\boldsymbol{\Sigma}_t^*$ are defined by Equation \eqref{eq:mtlcm-mlkl-mean-var} but with the ground-truth linear transformation $\mathbf{A}_*$, ground-truth causal indicators $\mathbf{c}_t^*$ and ground-truth spurious coefficients $\boldsymbol{\gamma}_t^*$. Then, \textcolor{black}{the latent factors recovered by MTLCM are guaranteed to be strictly strongly identifiable.} 
\end{theorem}
\begin{remark}
    The proof of Theorem \ref{thm:strong-id} can be found in Appendix \ref{appendix:proof-causal-id}. The first part of the proof adapts the proof technique from \citet{Khemakhem_Kingma_Monti_Hyvarinen_2020} to show identifiability up to block permutations and scaling. \textcolor{black}{The second part of the proof is novel, which leverages the properties of the linear likelihood as shown in Equation \eqref{eq:s2-lkl} to further reduce the block-identifiable equivalence class to permutations and scaling of the actual ground-truth latent factors. This resolves the point-wise indeterminacies of \citet{Khemakhem_Kingma_Monti_Hyvarinen_2020,Lu_Wu_Hernandez-Lobato_Scholkopf_2022} as they are only identifiable up to block transformations.}
\end{remark}

\subsection{Discussion of Model Assumptions}\label{sec:assumptions}
This section discusses some of the main assumptions underlying our model and their implications.

Regarding causal associations, our model proposes that the correlations between latent factors and the regression targets for each task be modelled as a partitioning of causal and spurious influences. \textcolor{black}{We provide real-world motivating examples to justify this assumption in Appendix \ref{appendix:motivating-examples}.} One could in principle consider other cases; one where there is no correlation between a latent variable and the target, or one where the correlation between a latent variable and the target arises from a confounding variable. We note that the former case could be handled by the model by treating it as a causal variable with a regression weight of zero. In the latter case, this confounding variable would then itself be a latent variable with a causal association to the target, and thus would not be unobserved. \textcolor{black}{These possibilities are depicted graphically in appendix \ref{appendix: dag-MTLCM}}. While it may be interesting for future work to consider the potential pairwise structure between latent variables, the simplicity of our model's optimization arises from its conditionally factorized structure.

Regarding probabilistic assumptions, while our model requires certain Gaussianity assumptions, we note that the final latent representation obtained by our model is a simple transformation only of the arbitrarily non-linear latent representation obtained from the MTRN described in Section \ref{sec:stage-1}. Thus, the conditional Gaussian form of Equation \eqref{eq:lin-causal-prior} may be viewed as a standard prior as in VAEs \citep{kingma2013auto}. The Gaussian assumption in Equation \eqref{eq:s2-lkl} follows the standard linear PCA model \citep{tipping1999probabilistic}, which is a natural choice given the linear identifiability result arising from the MTRN in Stage 1. The linear Gaussian regression model of Equation \eqref{eq: regression-model} is analogous to the standard predictive distribution for the final layer of regression neural networks. 

While the causal and probabilistic assumptions of our approach do not constitute the most general conceivable case, we note that there is an inherent tradeoff between full generality and tractability. Indeed, prior work which may theoretically allow for more general causal or probabilistic models typically entail an approximation in the optimization. Further, the empirical results on real-world data of Section \ref{sec:experiments} suggest that our approach may indeed be robust to moderate mis-specification.

\section{Experiments}\label{sec:experiments}
This section empirically validates our model's ability to recover canonical representations up to permutations and scaling for 
both synthetic and real-world data\footnote{Our code is available at \url{https://github.com/jdhorwood/mtlcm}.}.
We contrast our model with the more general identifiable models of iVAE \citep{Khemakhem_Kingma_Monti_Hyvarinen_2020} and iCaRL \citep{Lu_Wu_Hernandez-Lobato_Scholkopf_2022}. For a fair comparison, we also consider the multi-task extensions of iVAE and iCaRL, namely MT-iVAE and MT-iCaRL, which include the task variable $t$ in the conditioning variables $\mathbf{u}$ in their conditional priors $p_{\mathcal{T}}(\mathbf{z}|\mathbf{u})$, with the task-specific parameter $\mathcal{T}(t)=\{\mathbf{v}_t\}$ to be learned from data, which is the counterpart to $\mathcal{T}(t)=\{\mathbf{c}_t,\boldsymbol{\gamma}_t\}$ in our MTLCM but has no explicit interpretations with respect to a causal graph. We note that while the works of \citet{Fumero_Wenzel_Zancato_Achille_Rodolà_Soatto_Schölkopf_Locatello_2023, Lachapelle_Deleu_Mahajan_Mitliagkas_Bengio_Lacoste-Julien_Bertrand_2023} also consider methods for identifiability arising from learning across tasks, their approaches effectively implement a meta-learning setting (i.e., require that the support and query sets be disjoint in the bi-level optimization process). The assumption on task support variability for the latter also requires a much larger number of tasks (more than $10^5$ tasks as in their experiments) than what we consider here. These methods are thus not particularly well suited to comparison with our approach and we omit them from our baselines.
Detailed model configurations can be found in Appendix \ref{appendix: model-config}. Each experiment is run until convergence and repeated across 5 random seeds to guarantee reproducibility.

\begin{table}[t]
\centering
\caption{Identifiability performance for recovering the linearly transformed synthetic latent factors measured by strong MCC (\%).}
\resizebox{\textwidth}{!}{ 
\begin{tabular}{@{\hspace{0.01cm}}c@{\hspace{0.3cm}}c@{\hspace{0.3cm}}c@{\hspace{0.3cm}}c@{\hspace{0.3cm}}c@{\hspace{0.3cm}}c@{\hspace{0.3cm}}c@{\hspace{0.3cm}}c@{\hspace{0.3cm}}c@{\hspace{0.3cm}}c@{\hspace{0.01cm}}}
\toprule
\#Causal            & \multicolumn{5}{c}{2}       & \multicolumn{4}{c}{4}       \\ \cmidrule(lr){1-1} \cmidrule(lr){2-6} \cmidrule(lr){7-10}
\#Latent/Observed & 3/3 & 5/5 & 10/10 & 20/20 & 50/50 & 5/5 & 10/10 & 20/20 & 50/50 \\ \midrule
iVAE         & 87.75$\pm$5.02 & 78.02$\pm$0.73 & 81.36$\pm$0.57 & 82.30$\pm$0.27 & 81.96$\pm$0.07 & 81.67$\pm$2.97 & 74.29$\pm$0.30 & 77.57$\pm$0.15 & 79.79$\pm$0.10 \\
iCaRL        & 75.22$\pm$6.40 & 74.55$\pm$2.09 & 72.37$\pm$2.22 & 79.43$\pm$0.52 & 80.00$\pm$1.00 & 66.98$\pm$1.32 & 66.00$\pm$3.00 & 71.54$\pm$1.69 & 78.67$\pm$0.61 \\
MT-iVAE      & 91.78$\pm$8.12 & 90.14$\pm$5.01 & \textbf{99.89$\pm$0.04} & 97.90$\pm$1.51 & 90.56$\pm$3.18 & 76.09$\pm$7.69 & 76.36$\pm$2.32 & 98.42$\pm$0.88 & 94.53$\pm$2.49 \\
MT-iCaRL     & 81.09$\pm$3.37 & 71.12$\pm$2.97 & 76.13$\pm$0.53 & 79.26$\pm$1.00 & 81.30$\pm$0.84 & 61.55$\pm$1.26 & 64.04$\pm$1.08 & 72.79$\pm$1.92 & 79.54$\pm$0.59 \\
MTLCM        & \textbf{99.95$\pm$0.01} & \textbf{99.96$\pm$0.01} & 99.77$\pm$0.16 & \textbf{99.70$\pm$0.16} & \textbf{98.97$\pm$0.55} & \textbf{99.95$\pm$0.01} & \textbf{99.71$\pm$0.21} & \textbf{99.51$\pm$0.36} & \textbf{99.14$\pm$0.27}  \\ \bottomrule
\end{tabular}
}
\label{tab:linear-synthetic-comparison}
\end{table}
\begin{table}
\centering
\caption{Identifiability performance for recovering the non-linearly transformed synthetic latent factors measured by strong MCC (\%). The weak MCC (\%) for MTRN is also reported.}
\resizebox{\textwidth}{!}{ 
\begin{tabular}{@{\hspace{0.01cm}}c@{\hspace{0.3cm}}c@{\hspace{0.3cm}}c@{\hspace{0.3cm}}c@{\hspace{0.3cm}}c@{\hspace{0.3cm}}c@{\hspace{0.3cm}}c@{\hspace{0.3cm}}c@{\hspace{0.3cm}}c@{\hspace{0.3cm}}c@{\hspace{0.01cm}}}
\toprule
\#Causal & \multicolumn{3}{c}{4} & \multicolumn{3}{c}{8} & \multicolumn{3}{c}{12} \\ \cmidrule(lr){1-1} \cmidrule(lr){2-4} \cmidrule(lr){5-7} \cmidrule(lr){8-10}
\#Latent/Observed & 20/50 & 20/100 & 20/200 & 20/50 & 20/100 & 20/200 & 20/50 & 20/100 & 20/200 \\ \midrule
iVAE         & 73.11$\pm$1.13 & 77.42$\pm$0.20 & 76.95$\pm$0.31 & 65.18$\pm$1.49 & 68.66$\pm$0.14 & 69.05$\pm$0.17 & 58.70$\pm$0.33 & 60.33$\pm$0.27 & 59.85$\pm$0.31  \\
iCaRL        & 56.70$\pm$3.49 & 63.29$\pm$4.26 & 58.64$\pm$2.83 & 57.09$\pm$2.41 & 60.66$\pm$2.74 & 61.02$\pm$2.43 & 52.93$\pm$2.13 & 58.80$\pm$1.81 & 54.40$\pm$2.54  \\
MT-iVAE      & 71.78$\pm$1.45 & 80.14$\pm$0.37 & 73.89$\pm$2.98 & 65.44$\pm$1.60 & 69.31$\pm$0.35 & 68.56$\pm$0.34 & 55.79$\pm$1.61 & 60.56$\pm$0.23 & 59.61$\pm$0.30  \\
MT-iCaRL     & 67.57$\pm$1.97 & 70.26$\pm$3.22 & 65.52$\pm$0.65 & 63.37$\pm$0.84 & 63.75$\pm$2.19 & 61.61$\pm$1.52 & 57.13$\pm$1.07 & 60.56$\pm$0.15 & 58.10$\pm$1.04  \\
MTLCM        & \textbf{93.31$\pm$1.10} & \textbf{97.94$\pm$0.71} & \textbf{97.44$\pm$0.68} & \textbf{95.67$\pm$0.16} & \textbf{98.12$\pm$0.75} & \textbf{89.05$\pm$0.97} & \textbf{95.75$\pm$0.14} & \textbf{96.28$\pm$1.20} & \textbf{84.28$\pm$1.27}  \\
\midrule
MTRN (weak) & 89.38$\pm$0.71 & 96.15$\pm$0.91 & 96.19$\pm$0.87 & 93.96$\pm$0.22 & 97.63$\pm$0.79 & 87.75$\pm$0.99 & 95.14$\pm$0.17 & 96.12$\pm$1.27 & 83.70$\pm$1.22 \\
\bottomrule
\end{tabular}
\label{tab:nonlinear-synthetic-comparison}
}
\end{table}

\subsection{Synthetic Data}\label{sec:synthetic_exp}
We first validate our approach in the situation when the data generating process 
agrees with the assumptions of our models. For each task, we first sample the causal indicator variables $\mathbf{c}_t^*$. The causal latent factors $\mathbf{z}_c^*$ are then sampled from a standard Gaussian prior. These are then linearly combined according to random weights $\mathbf{w}_t^*$ to produce observed 
targets $y$ with a task-dependent noise corruption. Finally, the spurious variables $\mathbf{z}_s^*$ are generated via different weightings $\boldsymbol{\gamma}_t^*$ of the target $y$. This mirrors the causal data generating process described in Section \ref{sec:proposed_method}. For the linear case, we generate
observed data using random linear transformations of the ground-truth latent factors. 
For the non-linear case, we extend this to non-linear transformations parameterized by randomly initialized neural networks and demonstrate that our approach can be combined with 
the multi-task identifiability result up to linear transformations to recover permutations and scaling of the ground-truth. The detailed experimental setup can be found in Appendix \ref{appendix: experiment-settings-synth}.

\subsubsection{Linear Case}
We study the ability of our proposed multi-task linear causal model (MTLCM) to recover the latent factors up to permutations and scaling via the Mean Correlation Coefficient (MCC) as in \citet{Khemakhem_Kingma_Monti_Hyvarinen_2020}. The synthetic data is generated by sampling 200 tasks of 100 samples each. Each task
varies in its causal indicator variables $\mathbf{c}_t^*$, causal weights $\mathbf{w}_t^*$, and spurious coefficients $\boldsymbol{\gamma}_t^*$. We then transform the ground-truth latent factors $\mathbf{z}^*$ with a random invertible matrix $\mathbf{A}_*$ shared across all tasks to obtain linearly identifiable representations $\mathbf{h}$.
Identifiability in this setting is assessed by directly computing the MCC score between the representations obtained from our MTLCM and the ground-truth latent factors, which is referred to as strong MCC.
Since the data is known to be linearly identifiable, we use linear likelihoods for the baselines.

In Table \ref{tab:linear-synthetic-comparison}, we show that MTLCM manages to recover the ground-truth latent factors from $\mathbf{h}$ up to permutations and scaling, and the result is scalable as the number of latent factors and the number of causal factors increase. In contrast, iVAE, iCaRL and their multi-task extensions underperform our model by a large margin in most cases.
We find that for all tasks, the learned causal indicator variables also exactly match the ground-truth and the results from conditional independence testing \citep{chen2021causal,Lu_Wu_Hernandez-Lobato_Scholkopf_2022}. Ablation study for the effects of the learnable parameters and the type of linear transformation can be found in Appendix \ref{appendix:ablation study lin syn}.

\subsubsection{Non-Linear Case}
A more general analysis of the identifiability of our proposed approach is to consider the extension of the linear experiments to the setting of \textit{arbitrary} transformations of the latent factors. For this, we consider the case where random (non-linear) MLPs are used to transform $\mathbf{z}^*$ into higher dimensional observations $\mathbf{x}$.
By Corollary \ref{cor:lin-id-mtrn}, it is possible to recover linearly identifiable representations $\mathbf{h}$ of the data by training standard multi-task regression networks (MTRNs). Identifiability in this setting is assessed by first 
performing a Canonical Correlation Analysis (CCA) as in \citet{Roeder_Metz_Kingma_2021}, which linearly maps the obtained representations such that they maximize the covariance with the ground-truth latent factors. The resulting mapped representations can thus be 
compared with the ground-truth latent factors via the MCC score. 
This is referred to as weak MCC, which quantifies the linear identifiability of the learned representations from MTRNs.
We further train our MTLCM on the linearly identifiable representations $\mathbf{h}$ obtained from the MTRN to obtain identifiable representations up to permutations and scaling. Identifiability in this setting is assessed by directly computing the MCC score between the representations obtained from our MTLCM and the ground-truth latent factors as in the linear case (i.e., strong MCC).
We assess this for various dimensionalities of the observed data
and for different settings of the causal variables, where we generate 500 tasks of 200 samples each to improve convergence of the multitask model. 
The MTRN and the likelihoods in the baselines are parameterized by one-hidden-layer MLPs.

In Table \ref{tab:nonlinear-synthetic-comparison}, we find that the strong MCC for our MTLCM is able to match the weak MCC for the MTRN.
In contrast, the strong MCC for iVAE, iCaRL and their multi-task extensions significantly underperform MTLCM. Again, we find that for all tasks, the learned causal indicator variables exactly match the ground-truth and the results from conditional independence testing \citep{chen2021causal,Lu_Wu_Hernandez-Lobato_Scholkopf_2022}.

\begin{table}[t]
\centering
\caption{Identifiability performance for the latent factors learned on the superconductivity dataset measured by strong MCC (\%). The weak MCC (\%) for MTRN is also reported. ``$-$'' indicates divergence of optimization during training.}
\begin{tabular}{cccccc}
\toprule
Latent dim & 5 & 10 & 20 & 40 & 80 \\ \midrule
iVAE         & 32.87$\pm$1.16 & 33.21$\pm$1.04 & 30.68$\pm$0.39 & 37.41$\pm$0.84 & 45.52$\pm$0.81 \\
iCaRL        & $-$ & 32.23$\pm$0.61 & 35.62$\pm$0.40 & 32.58$\pm$2.16 & 32.19$\pm$2.45 \\
MT-iVAE      & 35.58$\pm$1.48 & 33.54$\pm$0.80 & 31.68$\pm$0.32 & 35.14$\pm$0.82 & 44.49$\pm$0.96 \\
MT-iCaRL     & $-$ & $-$ & $-$ & $-$ & 42.26$\pm$2.33 \\
MTLCM        & \textbf{98.90$\pm$0.03} & \textbf{96.93$\pm$0.12} & \textbf{84.56$\pm$1.11} & \textbf{46.31$\pm$0.34} & \textbf{48.94$\pm$2.16} \\
\midrule
MTRN (weak) & 98.85$\pm$0.03 & 97.17$\pm$0.04 & 93.23$\pm$0.08 & 78.58$\pm$0.09 & 52.02$\pm$0.19 \\
\bottomrule
\end{tabular}
\label{tab:superconductivity-comparison}
\end{table}

\subsection{Real-World Data}
\begin{figure}[t]
    \centering
    \includegraphics[width=0.9\linewidth]{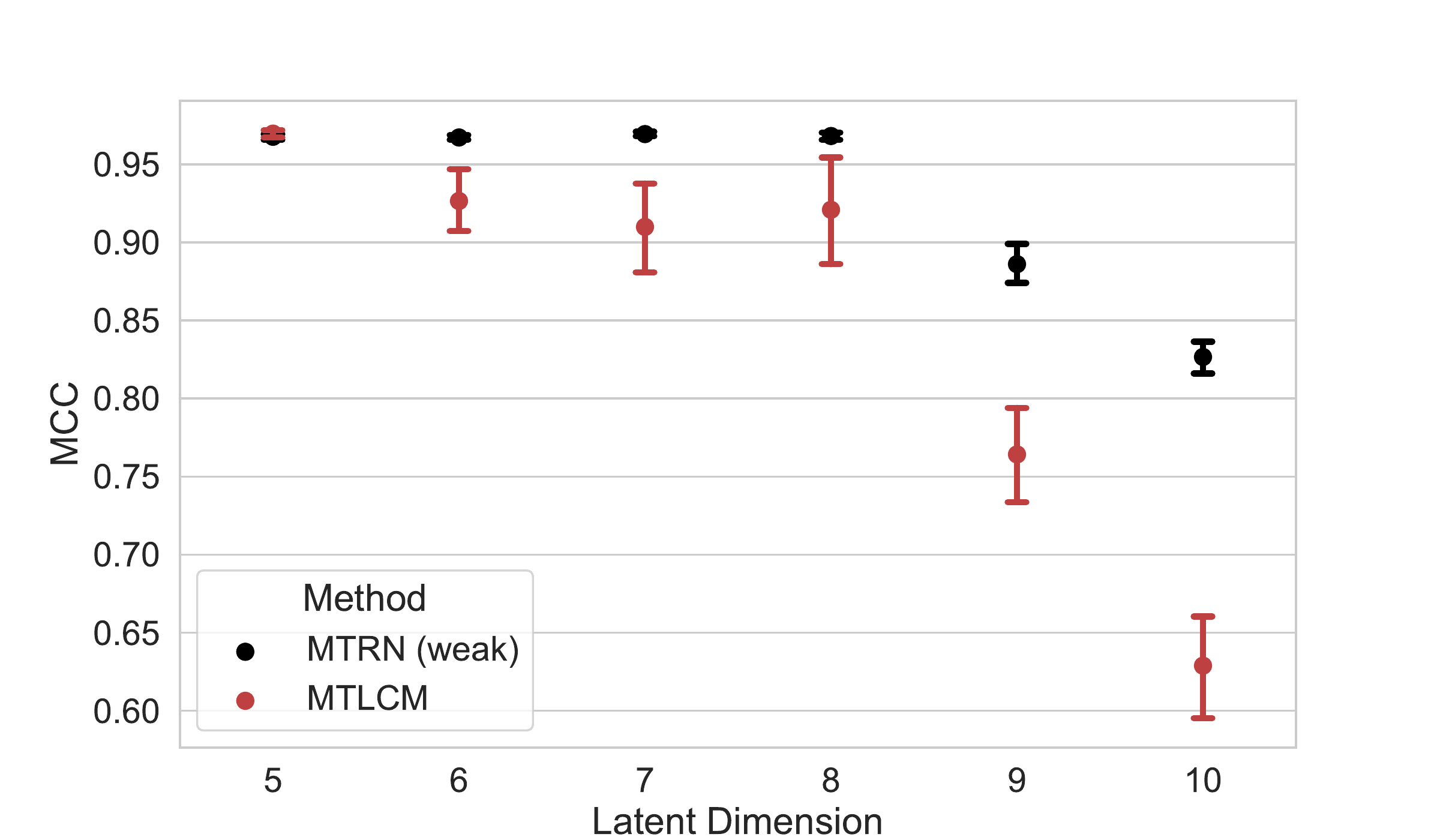}
    \caption{Identifiability performance for the latent factors learned on the QM9 dataset.}
    \label{fig:qm9-ident}
\end{figure}

We further evaluate our model on two real-world molecular datasets. We assume that the data $\mathbf{x}$ is generated by transforming some unknown ground-truth latent factors $\mathbf{z}^*$ with some unknown non-linear mixing function. Since $\mathbf{z}^*$ are unknown to us, identifiability in this setting is assessed by first training a model 5 times with different random seeds, then computing the MCC score between the data representations recovered by each pair of those 5 models, as in \citet{khemakhem2020ice}. As in Section \ref{sec:synthetic_exp}, we employ the weak MCC score to assess the linear identifiability of the representations $\mathbf{h}$ learned by the MTRN and the strong MCC score to assess the strictly strong identifiability of the latent factors $\mathbf{z}$ recovered by each method. \textcolor{black}{Given that the true latent dimension is unknown, we assess the identifiability of each model at gradually increasing latent dimensions. In practical scenarios, the latent dimension would be selected based on a similar model selection exercise. While the MCC threshold is likely to be dependent on the particular use case, the results in Table \ref{tab:superconductivity-comparison} and Figure \ref{fig:qm9-ident} suggest there is a relatively rapid shift from strong identifiability above 0.9 to much lower identifiability scores.}

\subsubsection{Superconductivity Dataset}
The superconductivity dataset \citep{hamidieh2018super} consists of $21,263$ superconductors. We consider the tasks of regressing $80$ readily computed target features such as mean atomic mass, thermal conductivity and valence of the superconductors from their chemical formulae, represented as discrete counts of the atoms present in the molecule. The MTRN and the likelihoods in the baselines are parameterized by MLPs.

In Table \ref{tab:superconductivity-comparison}, we find that the strong MCC for our MTLCM is greater than $0.96$ and is able to match the weak MCC for the MTRN when the dimensions of the latent representations are 5 and 10, showing that our method manages to recover canonical latent representations for the superconductors. Interestingly, the strong MCC score for the MTLCM decreases as we increase the number of latent factors in the model, suggesting that there are at most 10-20 independent tasks out of the 80 targets used for this data. In sharp contrast, all baseline models fail to recover identifiable latent factors for the superconductors in all cases as their strong MCC scores do not exceed $0.4$. There are several settings where optimization diverged during training, since VAE-based models are generally difficult to train on discrete inputs of chemical formulae.

\subsubsection{QM9 Dataset}
The QM9 dataset \citep{qm9, qm9_2} is a popular benchmark for molecular prediction tasks consisting of $134,000$ enumerated organic molecules of up to nine heavy atoms together with a set of $12$ calculated quantum chemical properties. 
In contrast to the more artificial superconductivity dataset, the QM9 dataset enables us to assess the feasibility of achieving identifiable representations in the context of highly non-trivial quantum chemical properties which are highly relevant to their pharmacological profile.
Accurately modeling this dataset requires us to capture potential three-dimensional atomic interactions, allowing us to assess the translation of our theoretical results to more complex equivariant graph neural network architectures.  For this reason, we use an equivariant graph neural network (EGNN) \citep{egnn} as the feature extractor for the MTRN. This enables the model to incorporate positional features of each atom while exhibiting equivariance to their rotation, translation or reflection. Given that the graph autoencoders proposed in \citet{egnn} and prior works \citep{Kipf_Welling_2016, Simonovsky_Komodakis_2018, Liu_Kumar_Ba_Kiros_Swersky_2019} do not provide a means of jointly decoding the feature and adjacency matrices, we do not consider the iVAE and iCaRL baselines for this dataset.

In Figure \ref{fig:qm9-ident}, the weak identifiability achieved from the MTRN implies that identifiability is achievable up to eight latent features, after which there is a sharp decline in MCC. \textcolor{black}{The implication of this observation is that there exist some redundancies between tasks (i.e., the number of effective tasks is less than the total number of tasks), which limit the maximal identifiable latent dimension. This is clearly the case for certain tasks. For example, prediction of the HOMO-LUMO gap can be directly obtained as a result of the difference between HOMO (highest occupied molecular orbital energy) and LUMO (lowest unoccupied molecular orbital energy) values.} Nonetheless, the MTLCM is able to closely approximate the weak MCC score up to eight latent factors, always surpassing a score of 0.9, demonstrating its ability to recover permutation identifiable representations in the context of realistic molecular datasets.

\section{Conclusion}\label{sec:conclusion}
We have proposed a novel perspective on the problem of identifiable representations by exploring the implications of explicitly modeling task structures. We have shown that this 
implies new identifiability results for linear equivalence classes in the general case of multi-task regression. Furthermore, while spurious correlations have been shown to be a
failure case of deep learning in many recent works, we have demonstrated that such latent spurious signals may in fact be leveraged to \textit{improve} the ability of a model to recover more robust disentangled representations (i.e., point-wise identifiability). In particular, when the latent space is explicitly represented as consisting of a partitioning of causal and spurious features per task, the linear identifiability result
of the multi-task setting may be reduced to identifiability up to simple permutations and scaling under sufficient variability conditions. We have thoroughly discussed the assumptions underlying our proposed model and their implications. Empirically, we have confirmed that the theoretical results hold for both linear and non-linear synthetic data and for two real-world molecular datasets of superconductors and organic small molecules.
We anticipate that this may reveal new research directions for the study of both causal representations and synergies with multi-task methods, and hope that these methods will enable robust generalization across tasks.

\subsubsection*{Acknowledgments}
We thank Jon Paul Janet and Dino Oglic for helpful discussions. WC acknowledges funding via a Cambridge Trust Scholarship (supported by the Cambridge Trust) and a Cambridge University Engineering Department Studentship (under grant G105682 NMZR/089 supported by Huawei R\&D UK). Julien H acknowledges funding via a Cambridge Center for AI in Medicine Studentship (supported by AstraZeneca). JMHL acknowledges support from a Turing AI Fellowship under grant EP/V023756/1.

Part of this work was performed using resources provided by the Cambridge Service for Data Driven Discovery (CSD3) operated by the University of Cambridge Research Computing Service (\url{www.csd3.cam.ac.uk}), provided by Dell EMC and Intel using Tier-2 funding from the Engineering and Physical Sciences Research Council (capital grant EP/T022159/1), and DiRAC funding from the Science and Technology Facilities Council (\url{www.dirac.ac.uk}).

\bibliography{main}
\bibliographystyle{tmlr}

\clearpage
\appendix

\section{Proof of Theorem \ref{thm:lin-id}}\label{appendix:proof-lin-id}
\begin{proof}
    By the assumption that the predictive likelihoods for the two sets of parameters $\boldsymbol{\theta}'$ and $\boldsymbol{\theta}$ are equal, we have
    \begin{align}
        p_{\boldsymbol{\theta}'}(y|\mathbf{x},t)&=p_{\boldsymbol{\theta}}(y|\mathbf{x},t),\quad\forall t,\mathbf{x},y,\\
        \implies\mathcal{N}(y|f_{\boldsymbol{\phi}',\mathbf{w}_t'}(\mathbf{x}),\sigma_{r,t}^2) &= \mathcal{N}(y|f_{\boldsymbol{\phi},\mathbf{w}_t}(\mathbf{x}),\sigma_{r,t}^2),\quad\forall t,\mathbf{x},y,\\
        \implies \mathcal{N}(y|\mathbf{h}_{\boldsymbol{\phi}'}(\mathbf{x})^{\text{T}}\mathbf{w}_{t}',\sigma_{r,t}^2) &= \mathcal{N}(y|\mathbf{h}_{\boldsymbol{\phi}}(\mathbf{x})^{\text{T}}\mathbf{w}_{t},\sigma_{r,t}^2),\quad\forall t,\mathbf{x},y.
    \end{align}
    This implies that the means of the two Gaussian likelihoods on both sides are identical:
    \begin{align}\label{eq:multitask equal mean}
        \mathbf{h}_{\boldsymbol{\phi}'}(\mathbf{x})^{\text{T}}\mathbf{w}_{t}' &= \mathbf{h}_{\boldsymbol{\phi}}(\mathbf{x})^{\text{T}}\mathbf{w}_t,\quad\forall t,\mathbf{x},y.
    \end{align}
    By the assumption that $\text{Span}(\text{Im}(\mathbf{h}_{\boldsymbol{\phi}}))=\mathbb{R}^d$, there exist $d$ inputs $\mathbf{x}_1,\cdots,\mathbf{x}_d$ such that the matrix $\mathbf{H}=[\mathbf{h}_{\boldsymbol{\phi}}(\mathbf{x}_1),\cdots,\mathbf{h}_{\boldsymbol{\phi}}(\mathbf{x}_d)]\in\mathbb{R}^{d\times d}$ is invertible. 
    By the assumption that there exist $d$ tasks $\{t_i\}_{i=1}^d$ such that the set of regression weights $\{\mathbf{w}_{t_i}\}_{i=1}^d$ are linearly independent, we construct an invertible matrix $\mathbf{W}=[\mathbf{w}_{t_1},\cdots,\mathbf{w}_{t_d}]\in\mathbb{R}^{d\times d}$.
    For $\mathbf{h}_{\boldsymbol{\phi}'}$, we similarly define $\mathbf{H}'=[\mathbf{h}_{\boldsymbol{\phi}'}(\mathbf{x}_1),\cdots,\mathbf{h}_{\boldsymbol{\phi}'}(\mathbf{x}_d)]\in\mathbb{R}^{d\times d}$ and $\mathbf{W}'=[\mathbf{w}_{t_1}',\cdots,\mathbf{w}_{t_d}']\in\mathbb{R}^{d\times d}$.
    
    Now, we evaluate Equation \eqref{eq:multitask equal mean} at the $d$ inputs $\mathbf{x}_1,\cdots,\mathbf{x}_d$ and $d$ tasks $t_1,\cdots,t_d$ defined above, which gives us the following linear equation:
    \begin{align}
        (\mathbf{H}')^{\text{T}}\mathbf{W}'=\mathbf{H}\mathbf{W}.
    \end{align}
    Since both $\mathbf{H}$ and $\mathbf{W}$ are invertible by assumption and the weight matrices $\mathbf{W}$ and $\mathbf{W}'$ do not depend on the input $\mathbf{x}$, the matrix $\mathbf{W}'$ must be invertible.

    Now, evaluating Equation \eqref{eq:multitask equal mean} at the $d$ tasks $t_1,\cdots,t_d$, we have
    \begin{align}
        (\mathbf{W}')^{\text{T}}\mathbf{h}_{\boldsymbol{\phi}'}(\mathbf{x})
        &=\mathbf{W}^{\text{T}}\mathbf{h}_{\boldsymbol{\phi}}(\mathbf{x}),\quad\forall\mathbf{x}\\
        \implies \mathbf{h}_{\boldsymbol{\phi}'}(\mathbf{x})
        &=(\mathbf{W}')^{-\text{T}}\mathbf{W}^{\text{T}}\mathbf{h}_{\boldsymbol{\phi}}(\mathbf{x}),\quad\forall\mathbf{x}\\
        \implies \mathbf{h}_{\boldsymbol{\phi}'}(\mathbf{x})
        &=\mathbf{A}\mathbf{h}_{\boldsymbol{\phi}}(\mathbf{x}),\quad\forall\mathbf{x}.
    \end{align}
    Note that we have shown that $\mathbf{A}\coloneqq(\mathbf{W}')^{-\text{T}}\mathbf{W}^{\text{T}}$ is invertible. This completes the proof.
\end{proof}

\section{Proof of Theorem \ref{thm:strong-id}}\label{appendix:proof-causal-id}
\begin{proof}
    Let $k\coloneqq2d$ and $\mathbf{u}\coloneqq[y,t]$. We first rewrite the density of the conditional prior in the exponential family form:
    \begin{align}
        p_{\mathcal{T}}(\mathbf{z}|\mathbf{u}) = Z(\mathbf{u})^{-1}\exp\left(\mathbf{T}(\mathbf{z})^{\text{T}}\boldsymbol{\eta}(\mathbf{u})\right),
    \end{align}
    where $Z(\mathbf{u})=(2\pi)^{d/2}|\boldsymbol{\Lambda}_t|^{0.5}\exp\left(-\frac{1}{2}\mathbf{a}_t^{\text{T}}\boldsymbol{\Lambda}_t\mathbf{a}_t\right)$ is the normalizing constant, $\mathbf{T}(\mathbf{z})=\left[\begin{smallmatrix}\mathbf{z} \\ \mathbf{z}\circ\mathbf{z}\end{smallmatrix}\right]\in\mathbb{R}^{k}$ are the sufficient statistics, and $\boldsymbol{\eta}(\mathbf{u})=\left[\begin{smallmatrix}\boldsymbol{\Lambda}_t^{-1}\mathbf{a}_t \\ -\frac{1}{2}\text{diag}(\boldsymbol{\Lambda}_t)\end{smallmatrix}\right]\in\mathbb{R}^{k}$ are the natural parameters. We also rewrite the likelihood $p_{\mathbf{A}}(\mathbf{h}|\mathbf{z})$ using the noise distribution $p_{\boldsymbol{\epsilon}_o}(\boldsymbol{\epsilon}_o)=\mathcal{N}(\boldsymbol{\epsilon}_o|\mathbf{0},\sigma_o^2\mathbf{I})$:
    \begin{align}
        p_{\mathbf{A}}(\mathbf{h}|\mathbf{z}) =\mathcal{N}(\mathbf{h}|\mathbf{A}\mathbf{z},\sigma_{o}^{2}\mathbf{I})=\mathcal{N}(\mathbf{h}-\mathbf{A}\mathbf{z}|\mathbf{0},\sigma_o^2\mathbf{I})=p_{\boldsymbol{\epsilon}_o}(\mathbf{h}-\mathbf{A}\mathbf{z}).
    \end{align}
    Let $\mathbf{A}_{*}$ be the ground-truth transformation matrix such that $\mathbf{z}^*=\mathbf{A}_{*}^{-1}\mathbf{h}$, and $\mathcal{T}_{*}(t)=\{\mathbf{c}_t^*,\boldsymbol{\gamma}_t^*\}$ the ground-truth task-specific variables associated with each task $t$. The proof starts off by using the fact that we have maximized the marginal likelihood \eqref{eq:s2-mlkl} of $\mathbf{A}$ and $\mathcal{T}$ for all tasks. This means that the marginal likelihoods of the two models are identical:
    \begin{align}\label{eq:equal-mlkl}
        p_{\mathbf{A},\mathcal{T}}(\mathbf{h}|\mathbf{u})=p_{\mathbf{A}_{*},\mathcal{T}_{*}}(\mathbf{h}|\mathbf{u}),\quad\forall\mathbf{h},\mathbf{u}.
    \end{align}
    The goal is to show that the latent factors $\mathbf{z}=\mathbf{A}^{-1}\mathbf{h}$ recovered by our model and the ground-truth latent factor $\mathbf{z}^*=\mathbf{A}_{*}^{-1}\mathbf{h}$ are identical up to permutations and scaling for all $\mathbf{h}$. 
    
    Starting from the equality of the two marginal likelihoods \eqref{eq:equal-mlkl}, we have
    \begin{align}
        p_{\mathbf{A},\mathcal{T}}(\mathbf{h}|\mathbf{u})&=p_{\mathbf{A}_{*},\mathcal{T}_*}(\mathbf{h}|\mathbf{u})\\   
        \iff \int p_{\mathbf{A}}(\mathbf{h}|\mathbf{z})p_{\mathcal{T}}(\mathbf{z}|\mathbf{u})d\mathbf{z} &= \int p_{\mathbf{A}_{*}}(\mathbf{h}|\mathbf{z})p_{\mathcal{T}_{*}}(\mathbf{z}|\mathbf{u})d\mathbf{z}\\
        \iff \int p_{\boldsymbol{\epsilon}_o}(\mathbf{h}-\mathbf{A}\mathbf{z})p_{\mathcal{T}}(\mathbf{z}|\mathbf{u})d\mathbf{z} &= \int p_{\boldsymbol{\epsilon}_o}(\mathbf{h}-\mathbf{A}_{*}\mathbf{z})p_{\mathcal{T}_{*}}(\mathbf{z}|\mathbf{u})d\mathbf{z}\\
        \iff \int p_{\boldsymbol{\epsilon}_o}(\mathbf{h}-\Bar{\mathbf{h}})p_{\mathcal{T}}(\mathbf{A}^{-1}\Bar{\mathbf{h}}|\mathbf{u})\det(\mathbf{A})^{-1}d\Bar{\mathbf{h}} &= \int p_{\boldsymbol{\epsilon}_o}(\mathbf{h}-\Hat{\mathbf{h}})p_{\mathcal{T}_{*}}(\mathbf{A}_{*}^{-1}\Hat{\mathbf{h}}|\mathbf{u})\det(\mathbf{A}_{*})^{-1}d\Hat{\mathbf{h}}\label{eq:reduce-1}\\
        \iff \int p_{\boldsymbol{\epsilon}_o}(\mathbf{h}-\Bar{\mathbf{h}})\Tilde{p}_{\mathbf{A},\mathcal{T},\mathbf{u}}(\Bar{\mathbf{h}})d\Bar{\mathbf{h}} &= \int p_{\boldsymbol{\epsilon}_o}(\mathbf{h}-\Hat{\mathbf{h}})\Tilde{p}_{\mathbf{A}_{*},\mathcal{T}_{*},\mathbf{u}}(\Hat{\mathbf{h}})d\Hat{\mathbf{h}}\label{eq:reduce-2}\\
        \iff (p_{\boldsymbol{\epsilon}_o}*\Tilde{p}_{\mathbf{A},\mathcal{T},\mathbf{u}})(\mathbf{h})&=(p_{\boldsymbol{\epsilon}_o}*\Tilde{p}_{\mathbf{A}_{*},\mathcal{T}_{*},\mathbf{u}})(\mathbf{h})\label{eq:reduce-3}\\
        \iff F[p_{\boldsymbol{\epsilon}_o}](\boldsymbol{\omega})F[\Tilde{p}_{\mathbf{A},\mathcal{T},\mathbf{u}}](\boldsymbol{\omega})&=F[p_{\boldsymbol{\epsilon}_o}](\boldsymbol{\omega})F[\Tilde{p}_{\mathbf{A}_{*},\mathcal{T}_{*},\mathbf{u}}](\boldsymbol{\omega})\label{eq:reduce-4}\\
        \iff F[\Tilde{p}_{\mathbf{A},\mathcal{T},\mathbf{u}}](\boldsymbol{\omega})&=F[\Tilde{p}_{\mathbf{A}_{*},\mathcal{T}_{*},\mathbf{u}}](\boldsymbol{\omega})\label{eq:reduce-5}\\
        \iff \Tilde{p}_{\mathbf{A},\mathcal{T},\mathbf{u}}(\mathbf{h}) &= \Tilde{p}_{\mathbf{A}_{*},\mathcal{T}_{*},\mathbf{u}}(\mathbf{h})\\
        \iff p_{\mathcal{T}}(\mathbf{A}^{-1}\mathbf{h}|\mathbf{u})\det(\mathbf{A})^{-1} &= p_{\mathcal{T}_{*}}(\mathbf{A}_{*}^{-1}\mathbf{h}|\mathbf{u})\det(\mathbf{A}_{*})^{-1}\\
        \iff \mathbf{T}(\mathbf{A}^{-1}\mathbf{h})^{\text{T}}\boldsymbol{\eta}(\mathbf{u})-\log Z(\mathbf{u})-\log\det(\mathbf{A}) &= \mathbf{T}(\mathbf{A}_{*}^{-1}\mathbf{h})^{\text{T}}\boldsymbol{\eta}_{*}(\mathbf{u})-\log Z_*(\mathbf{u})-\log\det(\mathbf{A}_{*}),\label{eq:reduce-final}
    \end{align}
    where
    \begin{itemize}
        \item Equation \eqref{eq:reduce-1} follows by the definition $\Bar{\mathbf{h}}\coloneqq\mathbf{A}\mathbf{z},\Hat{\mathbf{h}}\coloneqq\mathbf{A}_{*}\mathbf{z}$,
        \item Equation \eqref{eq:reduce-2} follows by the definition $\Tilde{p}_{\mathbf{A},\mathcal{T},\mathbf{u}}(\Bar{\mathbf{h}})\coloneqq p_{\mathcal{T}}(\mathbf{A}^{-1}\Bar{\mathbf{h}}|\mathbf{u})\det(\mathbf{A})^{-1}$,
        \item $*$ in Equation \eqref{eq:reduce-3} denotes the convolution operator,
        \item $F$ in Equation \eqref{eq:reduce-4} denotes the Fourier transform operator,
        \item Equation \eqref{eq:reduce-5} follows since the characteristic function $F[p_{\boldsymbol{\epsilon}_o}]$ of the Gaussian noise $\boldsymbol{\epsilon}_o$ is nonzero almost everywhere.
    \end{itemize}
    Now we evaluate Equation \eqref{eq:reduce-final} at $\mathbf{u}=\mathbf{u}_0,\mathbf{u}_1,\cdots,\mathbf{u}_k$ from our assumption to obtain $k+1$ such equations and subtract the first equation from the remaining $k$ equations to obtain the following $k$ equations:
    \begin{align}
        \mathbf{T}(\mathbf{A}^{-1}\mathbf{h})^{\text{T}}(\boldsymbol{\eta}(\mathbf{u}_l)-\boldsymbol{\eta}(\mathbf{u}_0))+\log \frac{Z(\mathbf{u}_0)}{Z(\mathbf{u}_l)} &= \mathbf{T}(\mathbf{A}_{*}^{-1}\mathbf{h})^{\text{T}}(\boldsymbol{\eta}_{*}(\mathbf{u}_l)-\boldsymbol{\eta}_{*}(\mathbf{u}_0))+\log \frac{Z_*(\mathbf{u}_0)}{Z_*(\mathbf{u}_l)},
    \end{align}
    where $l=1,\cdots,k$. Putting those $k$ equations in the matrix-vector form gives
    \begin{align}
        \mathbf{L}^{\text{T}}\mathbf{T}(\mathbf{A}^{-1}\mathbf{h})=\mathbf{L}_{*}^{\text{T}}\mathbf{T}(\mathbf{A}_{*}^{-1}\mathbf{h}) +\mathbf{q},\label{eq:ivae-lin-wip}
    \end{align}
    where $q_l=\log \frac{Z_*(\mathbf{u}_0)Z(\mathbf{u}_l)}{Z_*(\mathbf{u}_l)Z(\mathbf{u}_0)}$, $\mathbf{L}$ is the invertible matrix defined in the assumption, and $\mathbf{L}_{*}$ is similarly defined for the second model. Since $\mathbf{L}$ is invertible, we can left multiply Equation \eqref{eq:ivae-lin-wip} by $\mathbf{L}^{-\text{T}}$ to obtain
    \begin{align}
        \mathbf{T}(\mathbf{A}^{-1}\mathbf{h})=\mathbf{M}\mathbf{T}(\mathbf{A}_{*}^{-1}\mathbf{h}) +\mathbf{r},\label{eq:ivae-lin-wip-2}
    \end{align}
    where $\mathbf{M}=\mathbf{L}^{-\text{T}}\mathbf{L}_{*}^{\text{T}}$ and $\mathbf{r}=\mathbf{L}^{-\text{T}}\mathbf{q}$. We note that our assumption only says $\mathbf{L}$ is invertible and tells us nothing about $\mathbf{L}_{*}$. Therefore, we need to show that $\mathbf{M}$ is invertible.
    Let $\mathbf{h}_l\coloneqq\mathbf{A}\mathbf{z}_l,~l=0,\cdots,k$. We evaluate Equation \eqref{eq:ivae-lin-wip-2} at these $k+1$ points to obtain $k+1$ such equations, and subtract the first equation from the remaining $k$ equations. This gives us
    \begin{equation}
        [\mathbf{T}(\mathbf{z}_1)-\mathbf{T}(\mathbf{z}_0),\cdots,\mathbf{T}(\mathbf{z}_k)-\mathbf{T}(\mathbf{z}_0)]=\mathbf{M}[\mathbf{T}(\mathbf{A}_{*}^{-1}\mathbf{h}_1)-\mathbf{T}(\mathbf{A}_{*}^{-1}\mathbf{h}_0),\cdots,\mathbf{T}(\mathbf{A}_{*}^{-1}\mathbf{h}_k)-\mathbf{T}(\mathbf{A}_{*}^{-1}\mathbf{h}_0)].\label{eq:ivae-lin-wip-3}
    \end{equation}
    We denote Equation \eqref{eq:ivae-lin-wip-3} by $\mathbf{R}\coloneqq\mathbf{M}\mathbf{R}_{*}$. We need to show that for any given $\mathbf{z}_0$, there exist $k$ points $\mathbf{z}_1,\cdots,\mathbf{z}_k$ such that the columns of $\mathbf{R}$ are linearly independent. Suppose, for contradiction, that the columns of $\mathbf{R}$ would never be linearly independent for any $\mathbf{z}_1,\cdots,\mathbf{z}_k$. Then the function $\mathbf{g}(\mathbf{z})\coloneqq\mathbf{T}(\mathbf{z})-\mathbf{T}(\mathbf{z}_0)$ would live in a $k-1$ or lower dimensional subspace, and therefore we would be able to find a non-zero vector $\boldsymbol{\lambda}\in\mathbb{R}^k$ orthogonal to that subspace. This would imply that $(\mathbf{T}(\mathbf{z})-\mathbf{T}(\mathbf{z}_0))^{\text{T}}\boldsymbol{\lambda}=\mathbf{0}$ and thus $\mathbf{T}(\mathbf{z})^{\text{T}}\boldsymbol{\lambda}=\mathbf{T}(\mathbf{z}_0)^{\text{T}}\boldsymbol{\lambda}=const,~\forall\mathbf{z}$, which contradicts the fact that our conditionally factorized multivariate Gaussian prior $p_{\mathcal{T}}(\mathbf{z}|\mathbf{u})$ is strongly exponential (see \citet{Khemakhem_Kingma_Monti_Hyvarinen_2020} for the definition). This shows that there exist $k$ points $\mathbf{z}_1,\cdots,\mathbf{z}_k$ such that the columns of $\mathbf{R}$ are linearly independent for any given $\mathbf{z}_0$. Therefore, $\mathbf{R}$ is invertible. Since $\mathbf{R}=\mathbf{M}\mathbf{R}_{*}$ and $\mathbf{M}$ is not a function of $\mathbf{z}$, this tells us that $\mathbf{M}$ must be invertible.
    
    Now that we have shown that $\mathbf{M}$ is invertible, the next step is to show that $\mathbf{M}$ is a block transformation matrix. We define a linear function $\mathbf{l}(\mathbf{z})=\mathbf{A}_{*}^{-1}\mathbf{A}\mathbf{z}$. Now, Equation \eqref{eq:ivae-lin-wip-2} becomes
    \begin{align}
        \mathbf{T}(\mathbf{z})=\mathbf{M}\mathbf{T}(\mathbf{l}(\mathbf{z})) +\mathbf{r}.\label{eq:identifiability}
    \end{align}
    We first show that the linear function $\mathbf{l}$ is a point-wise function. We differentiate both sides of the above equation w.r.t. $z_s$ and $z_t$ ($\forall s\not=t$) to obtain:
    \begin{align}
        \frac{\partial\mathbf{T}(\mathbf{z})}{\partial z_s} &= \mathbf{M}\sum_{i=1}^d\frac{\partial \mathbf{T}(\mathbf{l}(\mathbf{z}))}{\partial l_i(\mathbf{z})}\frac{\partial l_i(\mathbf{z})}{\partial z_s},\\
        \frac{\partial^2\mathbf{T}(\mathbf{z})}{\partial z_s\partial z_t} &= \mathbf{M}\sum_{i=1}^d \sum_{j=1}^d \frac{\partial^2 \mathbf{T}(\mathbf{l}(\mathbf{z}))}{\partial l_i(\mathbf{z}) \partial l_j(\mathbf{z})}\frac{\partial l_j(\mathbf{z})}{\partial z_t} \frac{\partial l_i(\mathbf{z})}{\partial z_s}  + \mathbf{M}\sum_{i=1}^d\frac{\partial \mathbf{T}(\mathbf{l}(\mathbf{z}))}{\partial l_i(\mathbf{z})}\frac{\partial^2 l_i(\mathbf{z})}{\partial z_s \partial z_t}.
    \end{align}
    Since the prior $p_{\mathcal{T}}(\mathbf{z}|\mathbf{u})$ is conditionally factorized, the second-order cross derivatives of the sufficient statistics are zeros. Therefore, the second equation above can be simplified as follows:
    \begin{align}
        \mathbf{0} &= \frac{\partial^2\mathbf{T}(\mathbf{z})}{\partial z_s\partial z_t}\\
        &= \mathbf{M}\sum_{i=1}^d \frac{\partial^2 \mathbf{T}(\mathbf{l}(\mathbf{z}))}{\partial l_i(\mathbf{z})^2}\frac{\partial l_i(\mathbf{z})}{\partial z_t} \frac{\partial l_i(\mathbf{z})}{\partial z_s}  + \mathbf{M}\sum_{i=1}^d\frac{\partial \mathbf{T}(\mathbf{l}(\mathbf{z}))}{\partial l_i(\mathbf{z})}\frac{\partial^2 l_i(\mathbf{z})}{\partial z_s \partial z_t}\\
        &= \mathbf{M}\mathbf{T}''(\mathbf{z})\mathbf{l}_{s,z}'(\mathbf{z})+\mathbf{M}\mathbf{T}'(\mathbf{z})\mathbf{l}_{s,z}''(\mathbf{z})\\
        &=\mathbf{M}\mathbf{T}'''(\mathbf{z})\mathbf{l}_{s,z}'''(\mathbf{z}),
    \end{align}
    where
    \begin{align}
        \mathbf{T}''(\mathbf{z})&=\left[\frac{\partial^2 \mathbf{T}(\mathbf{l}(\mathbf{z}))}{\partial l_1(\mathbf{z})^2},\cdots,\frac{\partial^2 \mathbf{T}(\mathbf{l}(\mathbf{z}))}{\partial l_d(\mathbf{z})^2}\right]\in\mathbb{R}^{k\times d},\\
        \mathbf{l}_{s,z}'(\mathbf{z})&=\left[\frac{\partial l_1(\mathbf{z})}{\partial z_t} \frac{\partial l_1(\mathbf{z})}{\partial z_s},\cdots,\frac{\partial l_d(\mathbf{z})}{\partial z_t} \frac{\partial l_d(\mathbf{z})}{\partial z_s}\right]^{\text{T}}\in\mathbb{R}^{d},\\
        \mathbf{T}'(\mathbf{z})&=\left[\frac{\partial \mathbf{T}(\mathbf{l}(\mathbf{z}))}{\partial l_1(\mathbf{z})},\cdots,\frac{\partial \mathbf{T}(\mathbf{l}(\mathbf{z}))}{\partial l_d(\mathbf{z})}\right]\in\mathbb{R}^{k\times d},\\
        \mathbf{l}_{s,z}''(\mathbf{z})&=\left[\frac{\partial^2 l_1(\mathbf{z})}{\partial z_s \partial z_t},\cdots,\frac{\partial^2 l_d(\mathbf{z})}{\partial z_s \partial z_t}\right]^{\text{T}}\in\mathbb{R}^{d},\\
        \mathbf{T}'''(\mathbf{z})&=[\mathbf{T}''(\mathbf{z}),\mathbf{T}'(\mathbf{z})]\in\mathbb{R}^{k\times k},\\
        \mathbf{l}_{s,z}'''(\mathbf{z})&=[\mathbf{l}_{s,z}'(\mathbf{z})^{\text{T}},\mathbf{l}_{s,z}''(\mathbf{z})^{\text{T}}]^{\text{T}}\in\mathbb{R}^{k}.
    \end{align}
    By Lemma 5 in \citet{Khemakhem_Kingma_Monti_Hyvarinen_2020} and the fact that $k=2d$, we have that the rank of $\mathbf{T}'''(\mathbf{z})$ is $2d$ and thus it is invertible for all $\mathbf{z}$. Since $\mathbf{M}$ is also invertible, we have that $\mathbf{M}\mathbf{T}'''(\mathbf{z})$ is invertible. Since $\mathbf{M}\mathbf{T}'''(\mathbf{z})\mathbf{l}_{s,z}'''(\mathbf{z})=\mathbf{0}$, it must be that $\mathbf{l}_{s,z}'''(\mathbf{z})=\mathbf{0},~\forall\mathbf{z}$. In particular, this means that $\mathbf{l}_{s,z}'(\mathbf{z})=\mathbf{0},~\forall s\not=t$ for all $\mathbf{z}$, which shows that the linear function $\mathbf{l}(\mathbf{z})=\mathbf{A}_{*}^{-1}\mathbf{A}\mathbf{z}$ is a point-wise linear function.
    
    Now, we are ready to show that $\mathbf{M}$ is a block transformation matrix. Without loss of generality, we assume that the permutation in the point-wise linear function $\mathbf{l}$ is the identity. That is, $\mathbf{l}(\mathbf{z})=[l_1 z_1,\cdots,l_d z_d]^{\text{T}}$ for some linear univariate scalars $l_1,\cdots,l_d\in\mathbb{R}$. Since $\mathbf{A}$ and $\mathbf{A}_{*}$ are invertible, we have that $\mathbf{l}^{-1}(\mathbf{z})=[l_1^{-1} z_1,\cdots,l_d^{-1} z_d]^{\text{T}}$. Define
    \begin{align}
        \Bar{\mathbf{T}}(\mathbf{l}(\mathbf{z}))\coloneqq\mathbf{T}(\mathbf{l}(\mathbf{z}))+\mathbf{M}^{-1}\mathbf{r}
    \end{align}
    and plug it into Equation \eqref{eq:identifiability} gives:
    \begin{align}
        \mathbf{T}(\mathbf{z})=\mathbf{M}\Bar{\mathbf{T}}(\mathbf{l}(\mathbf{z})).\label{eq:ivae-strong-id-wip}
    \end{align}
    We then apply $\mathbf{l}^{-1}$ to $\mathbf{z}$ at both sides of the Equation \eqref{eq:ivae-strong-id-wip} to obtain
    \begin{align}
        \mathbf{T}(\mathbf{l}^{-1}(\mathbf{z}))=\mathbf{M}\Bar{\mathbf{T}}(\mathbf{z}).
    \end{align}
    Since $\mathbf{l}$ is a point-wise function, for a given $q\in\{1,\cdots,k\}$ we have that 
    \begin{align}\label{eq:proof_perm}
        0 = \frac{\partial \mathbf{T}(\mathbf{l}^{-1}(\mathbf{z}))_q}{\partial z_s}=\sum_{j=1}^k M_{q,j}\frac{\partial \Bar{\mathbf{T}}(\mathbf{z})_j}{\partial z_s},\quad\text{for any $s$ such that $q\not=s$ and $q\not=2s$}.
    \end{align}
    Since the entries in $\Bar{\mathbf{T}}(\mathbf{z})$ are linearly independent, it must be that $M_{q,j}=0$ for any $j$ such that $\frac{\partial \Bar{\mathbf{T}}(\mathbf{z})_j}{\partial z_s}\not=0$. This includes the entries $j$ in $\Bar{\mathbf{T}}(\mathbf{z})$ which depend on $z_s$ (i.e., $j=s$ and $j=2s$). Note that this holds true for any $s$ such that $q\not=s$ and $q\not=2s$. Therefore, when $q$ is the index of an entry in the sufficient statistics $\mathbf{T}$ that corresponds to $z_i$ (i.e., $q=i$ or $q=2i$, and $i\not=s$), the only possible non-zero $M_{q,j}$ for $j$ are the ones that map between $\mathbf{T}_i(z_i)$ and $\Bar{\mathbf{T}}_i(l_i(z_i))$, where $\mathbf{T}_i$ are the factors in $\mathbf{T}$ that depend on $z_i$ and $\Bar{\mathbf{T}}_i$ are similarly defined. This shows that $\mathbf{M}$ is a block transformation matrix for each block $[z_i,z_i^2]$ with scaling factor $l_i$. That is, the only possible nonzero element in $\mathbf{M}$ are $M_{i,i}$, $M_{i,2i}$, $M_{2i,i}$, and $M_{2i,2i}$ for all $i\in\{1,\cdots,d\}$.
    
    Furthermore, for any $i\in\{1,\cdots,d\}$ we have that
    \begin{align}
        l_i^{-1} &= \frac{\partial \mathbf{T}(\mathbf{l}^{-1}(\mathbf{z}))_i}{\partial z_i}=\sum_{j=1}^k M_{i,j}\frac{\partial \Bar{\mathbf{T}}(\mathbf{z})_j}{\partial z_i}=M_{i,i}+2M_{i,2i}z_i,\\
        2l_i^{-1}z_i &= \frac{\partial \mathbf{T}(\mathbf{l}^{-1}(\mathbf{z}))_{2i}}{\partial z_i}=\sum_{j=1}^k M_{2i,j}\frac{\partial \Bar{\mathbf{T}}(\mathbf{z})_j}{\partial z_i}=M_{2i,i}+2M_{2i,2i}z_i.
    \end{align}
    This implies that $M_{i,2i}=0$ and $M_{2i,i}=0$ for any $i\in\{1,\cdots,d\}$, and $M_{i,i}=l_i^{-1}$ for $i\in\{1,\cdots,k\}$, which reduces $\mathbf{M}$ from a block transformation matrix to a permutation and scaling matrix. In particular, this means that the latent factors $z_i$ are identifiable up to permutations and scaling, with the transformation matrix $\mathbf{P}\in\mathbb{R}^{d\times d}$ defined by the first $d$ rows and $d$ columns of $\mathbf{M}$:
    \begin{align}
        \mathbf{A}^{-1}\mathbf{h}=\mathbf{P}\mathbf{A}_*^{-1}\mathbf{h}+\mathbf{r}
        \quad\iff\quad 
        \mathbf{h}=\mathbf{A}\mathbf{P}(\mathbf{A}_*^{-1}\mathbf{h})+\mathbf{A}\mathbf{r}.
    \end{align}
    Since $\mathbf{h}$ is linearly identifiable by assumption, it must be that $\mathbf{A}\mathbf{r}=\mathbf{0}$ by Definition \ref{def:weak-id}. Since $\mathbf{A}$ is invertible by assumption, it must be that $\mathbf{r}=\mathbf{0}$. Therefore, we have
    \begin{align}
         \mathbf{A}^{-1}\mathbf{h}=\mathbf{P}\mathbf{A}_*^{-1}\mathbf{h}.
    \end{align}
    
    This completes the proof.
\end{proof}

\section{Derivation of the Factorized Structured Conditional Prior}\label{appendix:uninformative prior}
Since no prior knowledge is assumed for the task-specific regression weights $\mathbf{w}_t\in\mathbb{R}^d$, we put an uninformative prior over $\mathbf{w}_t\in\mathbb{R}^d$ for all tasks $t$:
\begin{align}
    p(\mathbf{w}_t)\propto 1.
\end{align}
\textcolor{black}{
This uninformative prior can be thought of as a Gaussian prior with infinite variance:
\begin{equation}
    p(\mathbf{w}_t)=\lim_{\tau\to\infty}q_{\tau}(\mathbf{w}_t),
\end{equation}
where $q_{\tau}(\mathbf{w}_t)=\mathcal{N}(\mathbf{w}_t|\mathbf{0},\tau^2\mathbf{I})$. We marginalize out $\mathbf{w}_t$ from $p_{\mathcal{T}}(y|\mathbf{z}_c,t) =\mathcal{N}(y|(\mathbf{w}_t\circ\mathbf{c}_t)^{\text{T}}\mathbf{z},\sigma_{p}^{2})$ under our uninformative prior over $\mathbf{w}_t$, which makes the marginal uninformative:
\begin{align}
    p_{\mathcal{T}}'(y|\mathbf{z}_c,t)
    &=\int p_{\mathcal{T}}(y|\mathbf{z}_c,t)p(\mathbf{w}_t)d\mathbf{w}_t \\
    &= \int p_{\mathcal{T}}(y|\mathbf{z}_c,t)\lim_{\tau\to\infty}q_{\tau}(\mathbf{w}_t)d\mathbf{w}_t \\
    &=\lim_{\tau\to\infty}\int p_{\mathcal{T}}(y|\mathbf{z}_c,t)q_{\tau}(\mathbf{w}_t)d\mathbf{w}_t\\
    &=\lim_{\tau\to\infty}\mathcal{N}(y|0,\tau^2(\mathbf{z}\circ\mathbf{c}_t)^{\text{T}}(\mathbf{z}\circ\mathbf{c}_t)+\sigma_p^2)\\
    &\propto 1.
\end{align}
This is known as an improper uniform distribution since it does not necessarily integrate to one. However, it is worth noting that the posterior $p_{\mathcal{T}}(\mathbf{z}|y,t)$ is still well-defined even $p_{\mathcal{T}}'(y|\mathbf{z}_c,t)$ is improper. To see this, we denote the improper uniform distribution by $p_{\mathcal{T}}'(y|\mathbf{z}_c,t)=C$ for some constant $C$. Then, we have
}
\begin{align}
    p_{\mathcal{T}}(\mathbf{z}|y,t)
    &=\frac{p_{\mathcal{T}}(\mathbf{z}_c|t)p_{\mathcal{T}}'(y|\mathbf{z}_c,t)p_{\mathcal{T}}(\mathbf{z}_s|y,t)}{\int p_{\mathcal{T}}(\mathbf{z}_c|t)p_{\mathcal{T}}'(y|\mathbf{z}_c,t)p_{\mathcal{T}}(\mathbf{z}_s|y,t)d\mathbf{z}_s d\mathbf{z}_c}\\
    &=\frac{p_{\mathcal{T}}(\mathbf{z}_c|t)p_{\mathcal{T}}(\mathbf{z}_s|y,t)}{\int p_{\mathcal{T}}(\mathbf{z}_c|t)p_{\mathcal{T}}(\mathbf{z}_s|y,t)d\mathbf{z}_s d\mathbf{z}_c}\\
    &=p_{\mathcal{T}}(\mathbf{z}_c|t)p_{\mathcal{T}}(\mathbf{z}_s|y,t).
\end{align}
Since $p_{\mathcal{T}}(\mathbf{z}_c|t)$ factorizes over the causal latent factors and $p_{\mathcal{T}}(\mathbf{z}_s|y,t)$ factorizes over the spurious latent factors, the structured conditional prior $p_{\mathcal{T}}(\mathbf{z}|y,t)$ factorizes over all latent factors $\mathbf{z}$.

\textcolor{black}{
Furthermore, we verify that the compact expressions for the mean and variance of $p_{\mathcal{T}}(\mathbf{z}|y,t)$ in Equation \eqref{eq:con_prior_mean_var} are correct. Recall that Equation \eqref{eq:causal_prior} tells us that
\begin{equation}
    p_{\mathcal{T}}(\mathbf{z}_c|t)=\mathcal{N}(\mathbf{z}_c|\mathbf{0},\mathbf{I}),
\end{equation}
and Equation \eqref{eq:spurious_prior} tells us that
\begin{equation}
    p_{\mathcal{T}}(\mathbf{z}_s|y,t)=\mathcal{N}(\mathbf{z}_s|y\boldsymbol{\gamma}_t,\sigma_s^2\mathbf{I}).
\end{equation}
Recall that the compact expressions given by Equation \eqref{eq:con_prior_mean_var} are
\begin{equation}
    \mathbf{a}_t\coloneqq y\boldsymbol{\gamma}_t\circ(1-\mathbf{c}_t),\quad
    \boldsymbol{\Lambda}_t\coloneqq \text{diag}(\sigma_{s}^{2}(1-\mathbf{c}_t)+\mathbf{c}_t).
\end{equation}
For any causal latent variable $z_i$, we have $c_{t,i}=1$ and therefore $a_{t,i}=0$ and $\Lambda_{t,i}=1$. 
For any spurious latent variable $z_j$, we have $c_{t,j}=0$ and therefore $a_{t,j}=y\gamma_{t,j}$ and $\Lambda_{t,j}=\sigma_s$.
This verifies that Equation \eqref{eq:con_prior_mean_var} is correct.
}

\textcolor{black}{
\section{Derivation of the Marginal Likelihood for MTLCM}\label{appendix:derivation-mlkl}
The marginal likelihood for MTRN given by Equation \eqref{eq:s2-mlkl} is
\begin{equation}
    p_{\boldsymbol{\psi}}(\mathbf{h}|y,t)=\int p_{\mathbf{A}}(\mathbf{h}|\mathbf{z})p_{\mathcal{T}}(\mathbf{z}|y,t)d\mathbf{z}
    =\mathcal{N}(\mathbf{h}|\boldsymbol{\mu}_t,\boldsymbol{\Sigma}_t),
\end{equation}
where $p_{\mathbf{A}}(\mathbf{h}|\mathbf{z})=\mathcal{N}(\mathbf{h}|\mathbf{A}\mathbf{z},\sigma_{o}^{2}\mathbf{I})$ and $p_{\mathcal{T}}(\mathbf{z}|y,t)=\mathcal{N}(\mathbf{z}|\mathbf{a}_t,\boldsymbol{\Lambda}_t)$.
Equivalently, we can rewrite the likelihood in the following form:
\begin{equation}
    \mathbf{h} = \mathbf{A}\mathbf{z} + \boldsymbol{\varepsilon},
\end{equation}
where $p(\boldsymbol{\varepsilon})=\mathcal{N}(\boldsymbol{\varepsilon}|\mathbf{0},\sigma_{o}^{2}\mathbf{I})$. Since both $p_{\mathbf{A}}(\mathbf{h}|\mathbf{z})$ and $p_{\mathcal{T}}(\mathbf{z}|y,t)$ are linear Gaussians, we can derive closed-form expression for the mean $\boldsymbol{\mu}_t$ and covariance $\boldsymbol{\Sigma}_t$ using moment matching:
\begin{equation}
    \boldsymbol{\mu}_t=\mathbb{E}_{p_{\mathcal{T}}(\mathbf{z}|y,t)}[\mathbf{h}]=\mathbf{A}\mathbb{E}_{p_{\mathcal{T}}(\mathbf{z}|y,t)}[\mathbf{z}]=\mathbf{A}\mathbf{a}_t=y\mathbf{A}(\boldsymbol{\gamma}_t\circ(1-\mathbf{c}_t)),
\end{equation}
\begin{equation}
    \boldsymbol{\Sigma}_t = \text{Var}_{p_{\mathcal{T}}(\mathbf{z}|y,t)}[\mathbf{h}]=\mathbf{A}\text{Var}_{p_{\mathcal{T}}(\mathbf{z}|y,t)}[\mathbf{z}]\mathbf{A}^{\text{T}} + \text{Var}_{\boldsymbol{\varepsilon}}[\boldsymbol{\varepsilon}]=\mathbf{A}\boldsymbol{\Lambda}_t\mathbf{A}^{\text{T}}+\sigma_{o}^{2}\mathbf{I}=\mathbf{A}\text{diag}(\sigma_{s}^{2}(1-\mathbf{c}_t)+\mathbf{c}_t)\mathbf{A}^{\text{T}}+\sigma_o^2\mathbf{I}.
\end{equation}
This verifies that Equation \eqref{eq:s2-mlkl} is correct.
}

\section{\textcolor{black}{Possible DAGs for the Generating Process on the Target Variable}}\label{appendix: dag-MTLCM}

\textcolor{black}{Possible structures for the latent factors generating the target $y$ are given in Figure \ref{fig: dags}.}

\begin{figure}[!h]
    \centering
    \begin{subfigure}[t]{0.3\textwidth}
        \centering
        \begin{tikzpicture}
            \node[draw, circle] (zi) {$z_i$};
            \node[draw, circle, below of=zi] (zj) {$z_j$};
            \node[draw, circle, below of=zj] (zk) {$z_k$};
            \node[draw, circle, right of=zj, xshift=1cm, fill=gray!30] (y) {$y$};
            
            \draw[->] (y) -- (zi);
            \draw[->] (zj) -- (y);
            \draw[->] (zk) -- (y);
        \end{tikzpicture}
        \caption{Standard case}
    \end{subfigure}
    \begin{subfigure}[t]{0.3\textwidth}
        \centering
        \begin{tikzpicture}
            \node[draw, circle, ] (zi) {$z_i$};
            \node[draw, circle, below of=zi] (zj) {$z_j$};
            \node[draw, circle, below of=zj] (zk) {$z_k$};
            \node[draw, circle, right of=zj, xshift=1cm, fill=gray!30] (y) {$y$};
            \draw[->, dashed, green] (zi) -- (y);
            \draw[->] (zj) -- (y);
            \draw[->] (zk) -- (y);
        \end{tikzpicture}
        \caption{Uncorrelated case}
    \end{subfigure}
    \begin{subfigure}[t]{0.3\textwidth}
        \centering
        \begin{tikzpicture}
            \node[draw, circle, ] (zi) {$z_i$};
            \node[draw,  circle, left of=zj, xshift=-0.5cm] (c) {$c$};
            \node[draw, circle, below of=zi] (zj) {$z_j$};
            \node[draw, circle, below of=zj] (zk) {$z_k$};
            \node[draw, circle, right of=zj, xshift=1cm, fill=gray!30] (y) {$y$};
            \draw[->] (y) -- (zi);
            \draw[->] (zj) -- (y);
            \draw[->] (zk) -- (y);
            \draw[->, red] (c) -- (zj);
            \draw[->] (c) edge [bend right] (y);
        \end{tikzpicture}
        \caption{Confounder case}
    \end{subfigure}
    \caption{\textcolor{black}{Illustration of causal relationships which are captured by our model (a, b) and not captured by our model (c) for the relationships between latent variables and observed target $y$ for a given task. The red arrow in (c) indicates the portion of the graph which is not captured by MTLCM. Note than in (b), the existence of learned regression weights encapsulates this case if the learned weight is zero on the arrow $z_i \to y$. This is depicted with the dashed green arrow.}}
    \label{fig: dags}
\end{figure}

\section{\textcolor{black}{Motivating Examples for the Assumed Non-Causal Relationship}}\label{appendix:motivating-examples}
\textcolor{black}{
We acknowledge that indeed our assumed direct edge $y\to \mathbf{z}_s$ in Figure \ref{fig:causal-graph} does not in general capture all possible non-causal correlations between latent features and the target $y$, since the Reichenbach principle states that non-causal correlations can originate from either (1) a common cause (i.e., confounders) or (2) an anti-causal/spurious relationship (as assumed in this work).
}
\textcolor{black}{
However, we argue that there are many situations where the proposed model can be useful in practice, even if it does not explicitly model the confounders in full generality, since this anti-causal relationship (as in our paper) is well-documented in real-world examples in epidemiology and drug discovery. We provide a couple of real-world motivating examples to justify this anti-causal assumption below.}
\textcolor{black}{
\begin{itemize}
    \item \textbf{Epidemiology.} See Figure 1 in \citet{wang2021chain}, treating perceived pandemic impact or IES-R score as the regression target, or Figure 6 (right) in \citet{von2021simpson}, where testing status may well be included as a feature in estimating Case Fatality Rates, but there is likely to be causal influence between overall case fatality rate and testing policy. Broadly, any form of selection bias may lead to similar cases, where the selection criterion may be included in the feature set.
    \item \textbf{Drug discovery.} In most drug discovery campaigns, molecules to be tested are selected based on some structural similarities to an originally promising molecule (based on the quantity to be estimated, e.g. drug potency). Structural molecule features are then likely to be spuriously correlated with the regression target due to their selection criteria, without actually being involved in the drug’s mechanism of action.
\end{itemize}
}

\section{Model Configurations}\label{appendix: model-config}
In Stage 1, the learnable parameters of a multi-task regression network (MTRN) are the feature extractor parameters $\boldsymbol{\phi}$ and the task-specific regression weights $\mathbf{w}_t$ for all tasks $t$. These model parameters are learned by maximum likelihood as defined in Equation \eqref{eq:mtrn-loss}.

In Stage 2, the learnable parameters of a multi-task linear causal model (MTLCM) are the linear transformation $\mathbf{A}$, the causal indicators $\mathbf{c}_t$ for all tasks $t$, and the spurious coefficients $\boldsymbol{\gamma}_t$ for all tasks $t$. These are free parameters learned by maximum marginal likelihood as defined in Equation \eqref{eq:mtlcm-loss}. The binary causal indicators $\mathbf{c}_t$ are parameterized as free parameters squashed to $[0,1]$ by the sigmoid function. To allow for gradient update of $\mathbf{c}_t$, we do not binarize the output of the sigmoid function during training; instead, we use a soft version $\Tilde{\mathbf{c}}_t\in[0,1]^d$ during training. In practice, we find that this works well and all learned values for $c_{t,1}$ are very close to either 0 or 1. In the synthetic data setting, the learned causal indicators match the ground-truth values. In practice, we fix the spurious noise variance $\sigma_s$ and the observational noise variance $\sigma_0$ to some constants using cross validation.

For a fair comparison, we also consider the multi-task extensions of iVAE and iCaRL, MT-iVAE and MT-iCaRL, which include the task variable $t$ in the conditioning variables $\mathbf{u}$ in their conditional priors $p_{\mathcal{T}}(\mathbf{z}|\mathbf{u})$, with the task-specific parameter $\mathcal{T}(t)=\{\mathbf{v}_t\}$ to be learned from data, which is the counterpart to $\mathcal{T}(t)=\{\mathbf{c}_t,\boldsymbol{\gamma}_t\}$ in our MTLCM but has no explicit interpretations with respect to a causal graph. We set $\text{dim}(\mathbf{v}_t)=\text{dim}(\mathbf{c}_t)+\text{dim}(\boldsymbol{\gamma}_t)$ to ensure the same degree of flexibility as our MTLCM. The task-specific parameters $\mathbf{v}_t$ are free parameters learned together with other parameters in these models by optimizing their variational/score matching objective.

\section{Experiment Settings for the Synthetic Data}
\label{appendix: experiment-settings-synth}
This section details the precise process for the data generation of the synthetic data for both the linear and non-linear experiments in Section \ref{sec:synthetic_exp}. Algorithm \ref{alg:data-gen} details the full data generation process, Table \ref{tab:experimental_settings_linear} details the experiment hyperparameters used in the linear setting and Table \ref{tab:experimental_settings_nonlinear} details the hyperparameters used in the non-linear setting. The transformation in the linear experiments corresponds to either the identity, an orthogonal or a random matrix of size $d\times d$, while in the non-linear experiments it corresponds to a randomly initialized neural network with the specified hidden dimensions and relu activations.

\begin{algorithm}[t]
    \begin{algorithmic}
        \REQUIRE the number of latent features $d$, the number of causal features $N_c$, the number of tasks $N_t$, the number of points per task $N_s$, Ground-truth transformation $\mathbf{F}$ (random invertible matrix or random MLP)

        \STATE Set $\sigma_s = 0.1$ and $\sigma_o=0.01$
        \FOR{each task t}
        
	    \STATE Sample $d$ binary causal feature indicators $I_1^t, I_2^t, \cdots, I_d^t$
        \STATE Sample $d$ weights $w_1^t,w_2^t,\cdots,w_d^t\sim\mathcal{U}(0,1)$
        \STATE Sample spurious coefficients $\gamma_j^t\sim U(-1, 1)$ for all $j$ such that $I_j^t=0$.

	\FOR{each data point $\mathbf{x}_i^t$ in this task $t$}
        \STATE Sample causal features $z_{i,j} \sim \mathcal{N}(0, \sigma_s^2)$ for all $j$ such that $I_j^t = 1$
        \STATE Sample $\sigma_p^t \sim \mathcal{U}(2, 3)$ 
	\STATE Obtain target y = $\sum_{j | I_j = 1} z_{i,j} + \epsilon_p^t, ~\epsilon_p^t\sim \mathcal{N}(0, (\sigma_p^t)^2)$
	\STATE Obtain spurious features $z_{i,j} = \gamma_j  y + \epsilon_{s,i,j},~\epsilon_{s,i,j}\sim\mathcal{N}(0,\sigma_s^2)$ for all $j$ such that $I_j = 0$
        \STATE Obtain observed features via the transformation $\mathbf{x}_i^t = \mathbf{F}(\mathbf{z}_i^t) + \boldsymbol{\epsilon}_{o,i}^{t},~ \boldsymbol{\epsilon}_{o,i}^{t}\sim\mathcal{N}(\mathbf{0},\sigma_o^2\mathbf{I})$
        \ENDFOR
        \ENDFOR
    \end{algorithmic}
    \caption{Pseudocode for the data generating process in the synthetic data experiments}
    \label{alg:data-gen}
\end{algorithm}

\section{Ablation Study for the Linear Synthetic Data}\label{appendix:ablation study lin syn}

\begin{figure}[t]
    \centering
    \includegraphics[width=\linewidth]{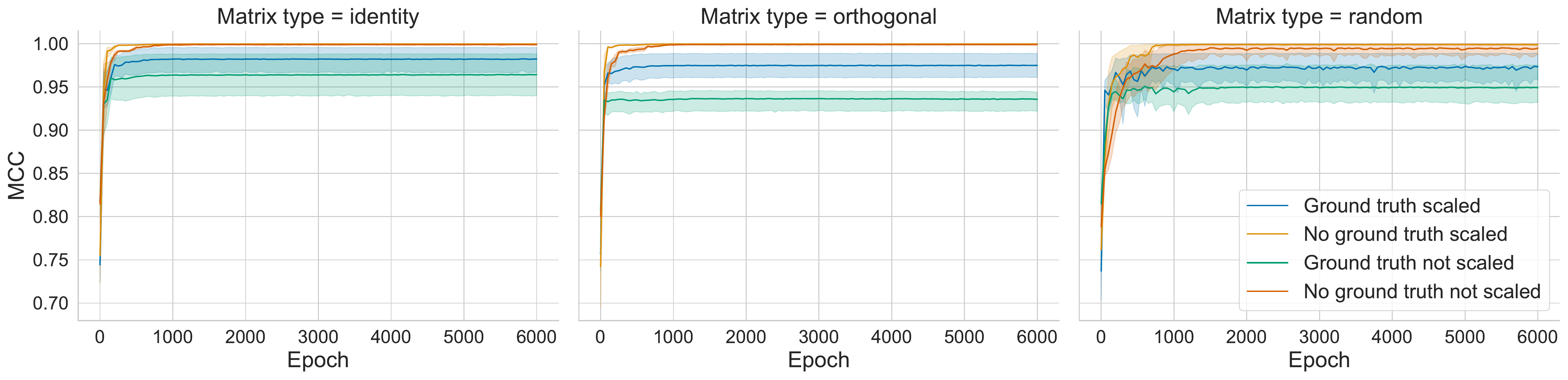}
    \caption{Convergence of the model in the case of transformations of the latent factors for identity, orthogonal and arbitrary linear transformations. Scaled means standardizing the features.}
    \label{fig:linear-synthetic-curves}
\end{figure}

In Figure \ref{fig:linear-synthetic-curves}, we contrast the 
effect of training only the linear transformation matrix $\mathbf{A}$ in our MTLCM when the ground-truth task variables $\mathbf{c}_t,\boldsymbol{\gamma}_t$ are known to the model, with the more general setting of learning all parameters jointly via maximum marginal likelihood. We assess the convergence of our multi-task linear causal model across 5 random seeds for increasingly complex 
linear transformations (identity, orthogonal, random) for data consisting of 10 latent factors with two causal features. Rather than
inhibiting convergence, we find that training all parameters jointly leads to improved performance, possibly due to additional flexibility in the parameterizations of the model. For all types of linear transformations, our model succeeds in recovering the ground-truth latent factors. In additional, we find that standardizing the features accelerates convergence..

\clearpage

\begin{table}[t]
    \caption{Experimental Settings for the Linear Synthetic Data}
    \label{tab:experimental_settings_linear}
    \centering
    \begin{tabular}{|c|c|}
    \hline
    Latent Dim & 3, 5, 10, 20, 50 \\ 
    \hline  
    Observation Dim & Latent Dim \\
    \hline
    Causal & 2, 4 \\ 
    \hline
    Seed & 1, 2, 3, 4, 5 \\ 
    \hline
    Matrix Type & random \\                                                
    \hline  
    \end{tabular}
\end{table} 

\begin{table}[t]
    \caption{Experimental Settings for the Non-Linear Synthetic Data}
    \label{tab:experimental_settings_nonlinear}
    \centering
    \begin{tabular}{|c|c|}
    \hline
    Observation Dim & 50, 100, 200 \\
    \hline
    Encoder Network Num Hidden Layer  & 1 \\
    \hline
    Encoder Network Hidden Dim & 2 * Observation dim \\
    \hline
    Latent Dim & 20 \\ 
    \hline  
    Num Causal & 4, 8, 12 \\ 
    \hline
    Seed & 1, 2, 3, 4, 5 \\      
    \hline  
    \end{tabular}
\end{table}

\end{document}